\documentclass[twoside,11pt]{article}
\usepackage{style/jair, rawfonts}
\ShortHeadings{Evaluation of Text Generation: A Survey}
{Celikyilmaz, Clark, \& Gao}
\firstpageno{1}

\usepackage{natbib}
\usepackage{paralist}
\usepackage[dvipsnames]{xcolor}
\usepackage{url}
\usepackage[pdftex]{graphicx}
\usepackage{latexsym}
\usepackage{hyperref}
\usepackage{amsmath}
\usepackage{amssymb}
\usepackage{array,multirow}
\usepackage{bbding}
\usepackage{pifont}
\usepackage{wasysym}
\usepackage{wrapfig}
\usepackage{subfigure}
\usepackage{caption}
\usepackage{mwe}
\usepackage{amsmath}
\usepackage{amssymb}
\usepackage{wasysym}
\usepackage{bbm}
\usepackage{CJKutf8}
\usepackage[colorinlistoftodos]{todonotes}
\usepackage[utf8]{inputenc}

\newcommand{\bleu}[1]{ \texttt{BLEU}}
\newcommand{\modelname}[1]{ \textsc{PlotMachines}}
\newcommand{\modelnameshort}[1]{\textsc{PM}}
\newcommand{\singlemem}[1]{\textsc{PM-Single}}
\newcommand{\dualmem}[1]{\textsc{PM-Full}}
\newcommand{\nomem}[1]{\textsc{PM-NoMem}}

\newcommand{\grover}[1]{ \textsc{Grover}}
\newcommand{\taskname}[1]{outline-conditioned story generation}
\makeatletter
\newcommand{\printfnsymbol}[1]{%
  \textsuperscript{\@fnsymbol{#1}}%
}
\makeatother

\title{Evaluation of Text Generation: A Survey}

\author{\name Asli Celikyilmaz \email aslic@fb.com \\
       \addr Facebook AI Research
       \AND
       \name Elizabeth Clark \email eaclark7@cs.washington.edu \\
       \addr University of Washington
       \AND
       \name Jianfeng Gao\email jfgao@microsoft.com \\
       \addr Microsoft Research
        }
       
\date{}

\begin{document}
\maketitle
\begin{abstract}
The paper surveys evaluation methods of natural language generation (NLG) systems that have been developed in the last few years. 
We group NLG evaluation methods into three categories: (1) human-centric evaluation metrics, (2) automatic metrics that require no training,
and (3) machine-learned metrics.
For each category, we discuss the progress that has been made and the challenges still being faced, with a focus on the evaluation of recently proposed NLG tasks and neural NLG models. We then present two examples for task-specific NLG evaluations for automatic text summarization and long text generation, and conclude the paper by proposing future research directions.\footnote{We are grateful to the following people: Rahul åJha, Sudha Rao, Ricky Loynd for their helpful comments and suggestions on earlier versions of this paper. We would like to thank the authors of the papers who gave us permission to use their figures, tables, and examples in our survey paper to summarize the related work. We would also like to thank the authors of the GEM Shared Task \citep{Gehrmann2021TheGB}, which aims to improve the evaluation of text generation models, for citing our work.
}
\end{abstract}

\section{Introduction}
\label{intro}
Natural language generation (NLG), a sub-field of natural language processing (NLP), deals with building software systems that can produce coherent and readable text \citep{buildingnlg}
NLG is commonly considered a general term which encompasses a wide range of tasks 
that take a form of input (e.g., a structured input like a dataset or a table, a natural language prompt or even an image) and output a sequence of text that is coherent and understandable by humans.
Hence, the field of NLG can be applied to a broad range of NLP tasks, such as generating responses to user questions in a chatbot, translating a sentence or a document from one language into another, offering suggestions to help write a story, or generating summaries of time-intensive data analysis.

The evaluation of NLG model output is challenging mainly because many NLG tasks are open-ended. 
For example, a dialog system can generate multiple plausible responses for the same user input. A document can be summarized in different ways.
Therefore, human evaluation remains the gold standard for almost all NLG tasks.
However, human evaluation is expensive, and researchers often resort to automatic metrics for quantifying day-to-day progress and for performing automatic system optimization.
Recent advancements in deep learning have yielded tremendous improvements in many NLP tasks.
This, in turn, presents a need for evaluating these deep neural network (DNN) models for NLG. 

In this paper we provide a comprehensive survey of NLG evaluation methods with a focus on evaluating neural NLG systems. We group evaluation methods into three categories: (1) human-centric evaluation metrics, (2) automatic metrics that require no training, and (3) machine-learned metrics.
For each category, we discuss the progress that has been made, the challenges still being faced, and proposals for new directions in NLG evaluation.

%######################################################
\subsection{Evolution of Natural Language Generation}
%######################################################
NLG is defined as the task of building software systems that can \textit{write} (e.g., producing explanations, summaries, narratives, etc.) in English and other human languages\footnote{From Ehud Reiter's Blog \citep{Ehudblog}.}. 
Just as people communicate ideas through writing or speech, NLG systems are designed to produce natural language text or speech that conveys ideas to its readers in a clear and useful way.
NLG systems have been used to generate text for many real-world applications, such as generating weather forecasts, carrying interactive conversations with humans in spoken dialog systems (chatbots), captioning images or visual scenes, translating text from one language to another, and generating stories and news articles.

NLG techniques range from simple template-based systems that generate natural language text using rules to machine-learned systems that have a complex understanding of human grammar\footnote{For an extensive survey on the evolution of NLG techniques, please refer to  \citet{Gatt2018SurveyOT}.}.
The first generation of automatic NLG systems uses rule-based or data-driven pipeline methods. In their book, \citet{nlgEhud} presented a classical three-stage NLG architecture.
The first stage is \textit{document planning}, which determines the content and its order and generates a text plan outlining the structure of messages. 
The second is the \textit{micro-planning} stage, when referring expressions that identify objects like entities or places are generated, along with the choice of words to be used and how they are aggregated. Collating similar sentences to improve readability with a natural flow also occurs in this stage. 
The last stage is \textit{realization}, in which 
% the abstract text specifications are converted to 
the actual text is generated, using linguistic knowledge about morphology, syntax, semantics, etc. 
Earlier work has focused on modeling discourse structures and learning representations of relations between text units for text generation \citep{mckeown85,marcu97,onoetal,dimlex},
for example using Rhetorical Structure Theory \citep{rst} or Segmented Discourse Representation Theory \citep{asher:lascarides:2003}. 
There is a large body of work that is based on template-based models and has used statistical methods to improve generation by introducing new methods such as sentence compression, reordering, lexical paraphrasing, and syntactic transformation, to name a few \citep{dsicchun,lsasumm,summstats,clarkecompression,quirkparaphrasing}.

These earlier text generation approaches and their extensions play an important role in the evolution of NLG research. Following this earlier work, 
an important direction that several NLG researchers have focused on is data-driven representation learning, which has gained attention with the availability of more data sources. 
Availability of large datasets, treebanks, corpora of referring expressions, as well as shared tasks have been beneficial in the progress of several NLG tasks today \citep{Gkatzia2015,Gatt07evaluatingalgorithms,Mairesse2010,Konstas2013,Konstas2012}. 

The last decade has witnessed a paradigm shift towards learning representations from large textual corpora in an unsupervised manner using deep neural network (DNN) models. Recent NLG models are built by training DNN models, typically on very large corpora of human-written texts. The paradigm shift starts with the use of recurrent neural networks \citep{rnn} (e.g., long short-term memory networks (LSTM)     \citep{Hochreiter1997LongSM}, gated recurrent units (GRUs) \citep{cho-etal-2014-learning}, etc.) for learning language representations, 
and later sequence-to-sequence learning \citep{s2s}, which opens up a new chapter characterised by the wide application of the encoder-decoder architecture. 
Although sequence-to-sequence models were originally developed for machine translation,
they were soon shown to improve performance across many NLG tasks. These models' weakness of capturing long-span dependencies in long word sequences motivated the development of \textit{attention networks} \citep{bahdanau2014neural} and \textit{pointer networks} \citep{vinyals2015pointer}.
The Transformer architecture \citep{vaswanietal}, which 
incorporates an encoder and a decoder, both implemented using the self-attention mechanism, is being adopted by new state-of-the-art NLG systems.
There has been a large body of research in recent years that focuses on improving the performance of NLG using large-scale pre-trained language models for contextual word embeddings
\citep{   elmo,bert,ernie,unilm},  using better sampling methods to reduce degeneration in decoding \citep{grover, nucleussampling},  and learning to generate tex t with better discourse structures and narrative flow \citep{planandwrite,storybyangela,pplm,plotmachines}. 

Neural models have been applied to many NLG tasks, which we will discuss in this paper, including:
\begin{itemize}
    \setlength\itemsep{-0.5em}
    \item summarization: common tasks include single or multi-document tasks, query-focused or generic summarization, and summarization of news, meetings, screen-plays, social blogs, etc. 
    \item machine translation: sentence- or document-level.
    \item dialog response generation: goal-oriented or chit-chat dialog.
    \item paraphrasing
    \item question generation
    \item long text generation: most common tasks are story, news, or poem generation.
    \item data-to-text generation: e.g., table summarization.
    \item caption generation from non-textual input: input can be tables, images, or sequences of video frames (e.g., in visual storytelling), to name a few. 
\end{itemize}

%#####################################################################
\subsection{Why a Survey on Evaluation of Natural Language Generation}
%#####################################################################

Text generation is a key component of language translation, chatbots, question answering, summarization, and several other applications that people interact with everyday. 
%From a model training perspective, 
Building language models using traditional approaches is a complicated task that needs to take into account multiple aspects of language, including linguistic structure, grammar, word usage, and reasoning, and thus requires non-trivial data labeling efforts.
Recently, Transformer-based neural language models have been shown to be very effective in leveraging large amounts of raw text corpora from online sources (such as Wikipedia, search results, blogs, Reddit posts, etc.). For example, one of most advanced neural language models, GPT-2/3 \citep{radford2019language,NEURIPS2020_1457c0d6}, can generate long texts that are almost indistinguishable from human-generated texts \citep{grover, NEURIPS2020_1457c0d6}. Empathetic social chatbots, such as XiaoIce \citep{zhou2020design},
%\footnote{https://www.msxiaobing.com/} 
seem to understand human dialog well and can generate interpersonal responses to establish long-term emotional connections with users.
%on an emotional level and engage in interpersonal and long-form communication. 

Many NLG surveys have been published in the last few years 
% to summarize recent approaches for text generation
\citep{Gatt2018SurveyOT,texygen,Zhang2019}. 
Others survey specific NLG tasks or NLG models, such as 
image captioning \citep{BernardiCEEEIKM16,kilickaya_2016,imagecapsurvey,imagecaptsurvey2019,imagecaptsurvey2018},
% \jianfeng{citation?} , 
machine translation \citep{mtsurvey1,mtsurvey2,maruf2019survey}, summarization \citep{evalsumm,shi2018}, question generation \citep{recentqg}, extractive key-phrase generation \citep{ano2019keyphrase}, deep generative models \citep{pelsmaeker2019,dlvm}, text-to-image synthesis \citep{agnese2019survey}, and dialog response generation \citep{liu-etal-2016-evaluate,novikova-etal-2017-need,surveydialogeval, dialogeval20192,Gao2018}, 
% document-level machine translation \citep{maruf2019survey}, 
to name a few.

There are only a few published papers that review evaluation methods for specific NLG tasks, such as image captioning \citep{kilickaya_2016}, machine translation \citep{mteval}, online review generation \citep{Garbacea2019}, interactive systems \citep{hastie-belz-2014-comparative}, and conversational dialog systems \citep{surveydialogeval}, and for human-centric evaluations \citep{humaneval1,humaneval2}. 
The closest to our paper is the NLG survey paper of \cite{gkatzia-mahamood-2015-snapshot}, which includes a section on NLG evaluation metrics. 

Different from this work, our survey is dedicated to NLG evaluation, with a focus on the evaluation metrics developed recently for neural text generation systems, and provides an in-depth analysis of existing metrics to-date. 
%We also provide additional task-specific evaluation metrics for providing a guide for evaluating long-text-generation from different aspects. 
To the best of our knowledge, our paper is the most extensive and up-to-date survey on NLG evaluation.

%########################################
\subsection{Outline of The Survey}
%########################################
We review NLG evaluation methods in three categories in Sections~\ref{human}-\ref{model}:
\begin{itemize}
\item \textbf{Human-Centric Evaluation.} The most natural way to evaluate the quality of a text generator is to involve \textit{humans as judges}. 
%which we will investigate in our first category of evaluation methods. 
Naive or expert subjects are asked to rate or compare texts generated by different NLG systems or to perform a Turing test \citep{turing} to distinguish machine-generated texts from human-generated texts. 

Some human evaluations may require the judging of task-specific criteria (e.g., evaluating that certain entity names appear correctly in the text, such as in health report summarization), while other human evaluation criteria can be generalized for most text generation tasks (e.g., evaluating the fluency or grammar of the generated text). 

\item \textbf{Untrained Automatic Metrics.} This category, also known as \textit{automatic metrics}, is the most commonly used in the research community. 
These evaluation methods compare machine-generated texts to human-generated texts (reference texts) based on the same input data and use metrics that do not require machine learning but are simply based on string overlap, content overlap, string distance, or lexical diversity, such as $n$-gram match and distributional similarity. 
For most NLG tasks, it is critical to select the right automatic metric that measures the aspects of the generated text that are consistent with the original design goals of the NLG system.

\item \textbf{Machine-Learned Metrics.} These metrics are often based on machine-learned models, which are used to measure the similarity between two machine-generated texts or between machine-generated and human-generated texts. These models can be viewed as digital judges that simulate human judges.
We investigate the differences among these evaluations and shed light on the potential factors that contribute to these differences. 
\end{itemize}
To see how these evaluation methods are applied in practice, we look at the role NLG shared tasks have played in NLG model evaluation (Section~\ref{sharedtasks}) and at how evaluation metrics are applied in two NLG subfields (Section~\ref{applications}): automatic document summarization and long-text generation.
Lastly, we conclude the paper with future research directions for NLG evaluation (Section~\ref{conclusion}).

\section{Human-Centric Evaluation Methods}
\label{human}
Whether a system is generating an answer to a user's query, a justification for a classification model's decision, or a short story, the ultimate goal in NLG is to generate text that is valuable to people. For this reason, human evaluations are typically viewed as the most important form of evaluation for NLG systems and are held as the gold standard when developing new automatic metrics. Since automatic metrics still fall short of replicating human decisions \citep{Reiter2009AnII,krahmer2010empirical,reiter-2018-structured}, many NLG papers include some form of human evaluation. For example, \citet{hashimoto-etal-2019-unifying} report that 20 out of 26 generation papers published at ACL2018 presented human evaluation results.

While human evaluations give the best insight into how well a model performs in a task, it is worth noting that human evaluations also pose several challenges. First, human evaluations can be expensive and time-consuming to run, especially for the tasks that require extensive domain expertise. While online crowd-sourcing platforms such as Amazon Mechanical Turk have enabled researchers to run experiments on a larger scale at a lower cost, they come with their own problems, such as maintaining quality control \citep{Ipeirotis2010QualityMO, Mitra2015ComparingPA}.
Second, even with a large group of annotators, there are some dimensions of generated text quality that are not well-suited to human evaluations, such as diversity \citep{hashimoto-etal-2019-unifying}.
Thirdly, there is a lack of consistency in how human evaluations are run, which prevents researchers from reproducing experiments and comparing results across systems. This inconsistency in evaluation methods is made worse by inconsistent reporting on methods; details on how the human evaluations were run are often incomplete or vague. For example, \citet{Lee2021HumanEO} find that in a sample of NLG papers from ACL and INLG, only 57\% of papers report the number of participants in their human evaluations.

In this section, we describe common approaches researchers take when evaluating generated text using only human judgments, grouped into intrinsic (\S\ref{sec:intrinsic}) and extrinsic (\S\ref{sec:extrinsic}) evaluations \citep{belz-reiter-2006-comparing}. 
However, there are other ways to incorporate human subjects into the evaluation process, such as training models on human judgments, which will be discussed in Section \ref{model}.

\subsection{Intrinsic Evaluation}
\label{sec:intrinsic}
An \textit{intrinsic evaluation} asks people to evaluate the quality of generated text, either overall or along some specific dimension (e.g., fluency, coherence, correctness, etc.). This is typically done by generating several samples of text from a model and asking human evaluators to score their quality.

\begin{figure}
  \vspace{-20pt}
\centering
\subfigure[Likert-scale question]{\label{fig:likert}\includegraphics[width=\textwidth,trim={0.5cm 9cm 1cm 3cm},clip]{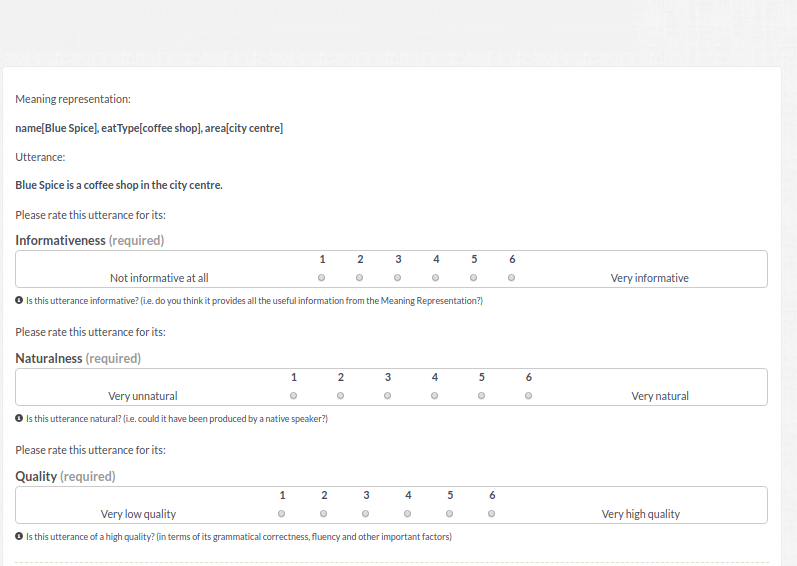}}
\subfigure[RankME-style question]{\label{fig:rankme}\includegraphics[width=\textwidth, trim={0 2cm 0 10cm},clip]{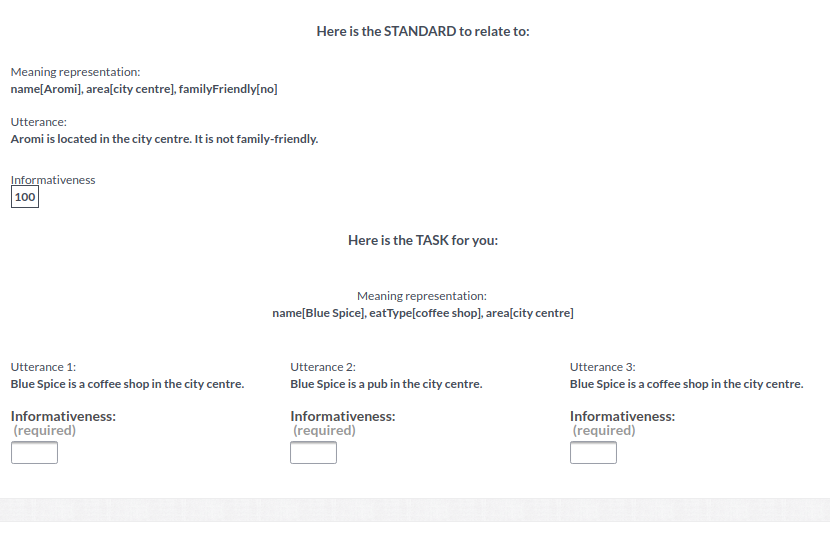}}
\caption{Two different methods for obtaining intrinsic evaluations of text generated from a meaning representation. Image Source: \citep{novikova-etal-2018-rankme}, \url{https://github.com/jeknov/RankME} 
}
  \vspace{-10pt}
\end{figure}

The simplest way to get this type of evaluation is to show the evaluators the generated texts one at a time and have them judge their quality individually. They are asked to vote whether the text is good or bad, or to make more fine-grained decisions by marking the quality along a Likert or sliding scale (see Figure \ref{fig:likert}). However, judgments in this format can be inconsistent and comparing these results is not straightforward; \citet{humaneval2} find that analysis on NLG evaluations in this format is often done incorrectly or with little justification for the chosen methods.

To more directly compare a model's output against baselines, model variants, or human-generated text, intrinsic evaluations can also be performed by having people choose which of two generated texts they prefer, or more generally, rank a set of generated texts. This comparative approach has been found to produce higher inter-annotator agreement \citep{callison-burch-etal-2007-meta} in some cases. However, while it captures models' relative quality, it does not give a sense of the absolute quality of the generated text. One way to address this is to use a method like RankME \citep{novikova-etal-2018-rankme}, which adds magnitude estimation \citep{10.2307/416793} to the ranking task, asking evaluators to indicate how much better their chosen text is over the alternative(s) (see Figure \ref{fig:rankme}). Comparison-based approaches can become prohibitively costly (by requiring lots of head-to-head comparisons) or complex (by requiring participants to rank long lists of output) when there are many models to compare, though there are methods to help in these cases. For example, best-worst scaling \citep{louviere_flynn_marley_2015} has been used in NLG tasks \citep{kiritchenko-mohammad-2016-capturing, KoncelKedziorski2019TextGF} to simplify comparative evaluations; best-worst scaling asks participants to choose the best and worst elements from a set of candidates, a simpler task than fully ranking the set that still provides reliable results.
 
Almost all text generation tasks today are evaluated with intrinsic human evaluations. Machine translation is one of the text generation tasks in which intrinsic human evaluations have made a huge impact on the development of more reliable and accurate translation systems, as automatic metrics are validated through correlation with human judgments.  One metric that is most commonly used to judge translated output by humans is measuring its \textit{adequacy}, which is defined by the \textit{Linguistic Data Consortium} as ``how much of the meaning expressed in the gold-standard translation or source is also expressed in the target translation''\footnote{https://catalog.ldc.upenn.edu/docs/LDC2003T17/TransAssess02.pdf}.
The annotators must be bilingual in both the source and target languages in order to judge whether the information is preserved across translation. Another dimension of text quality commonly considered in machine translation is \textit{fluency}, which measures the quality of the generated text only (e.g., the target translated sentence), without taking the source into account. It accounts for criteria such as grammar, spelling, choice of words, and style. A typical scale used to measure fluency is based on the question “Is the language in the output fluent?”. Fluency is also adopted in several text generation tasks including document summarization \citep{celikyilmaz2018deep,ranksummlapata18}, recipe generation \citep{bosselut2018discourseaware}, image captioning \citep{fluencyimagecaptioning}, video description generation \citep{videocvpr2018}, and question generation \citep{qglearningtoask}, to name a few. 

While fluency and adequacy have become standard dimensions of human evaluation for machine translation, not all text generation tasks have an established set of dimensions that researchers use. Nevertheless, there are several dimensions that are common in human evaluations for generated text. 
As with adequacy, many of these dimensions focus on the content of the generated text. \textit{Factuality} is important in tasks that require the generated text to accurately reflect facts described in the context. For example, in tasks like data-to-text generation or summarization, the information in the output should not contradict the information in the input data table or news article. This is a challenge to many neural NLG models, which are known to ``hallucinate'' information \citep{nucleussampling, welleck2019neural}; \citet{Maynez2020OnFA} find that over 70\% of generated single-sentence summaries contained hallucinations, a finding that held across several different modeling approaches. 
Even if there is no explicit set of facts to adhere to, researchers may want to know how well the generated text follows rules of \textit{commonsense} or how \textit{logical} it is. 
For generation tasks that involve extending a text, researchers may ask evaluators to gauge the \textit{coherence} or \textit{consistency} of a text---how well it fits the provided context. For example, in story generation, do the same characters appear throughout the generated text, and do the sequence of actions make sense given the plot so far?

Other dimensions focus not on what the generated text is saying, but how it is being said. As with fluency, these dimensions can often be evaluated without showing evaluators any context. This can be something as basic as checking for simple language errors by asking evaluators to rate how \textit{grammatical} the generated text is. It can also involve asking about the overall \textit{style}, \textit{formality}, or \textit{tone} of the generated text, which is particularly important in style-transfer tasks or in multi-task settings. \citet{hashimoto-etal-2019-unifying} ask evaluators about the \textit{typicality} of generated text; in other words, how often do you expect to see text that looks like this? These dimensions may also focus on how efficiently the generated text communicates its point by asking evaluators how \textit{repetitive} or \textit{redundant} it is.

Note that while these dimensions are common, they may be referred to by other names, explained to evaluators in different terms, or measured in different ways \citep{Lee2021HumanEO}. \citet{Howcroft2020TwentyYO} found that \~25\% of generation papers published in the last twenty years failed to mention what the evaluation dimensions were, and less than half included definitions of these dimensions. More consistency in how user evaluations are run, especially for well-defined generation tasks, would be useful for producing comparable results and for focused efforts for improving performance in a given generation task. One way to enforce this consistency is by handing over the task of human evaluation from the individual researchers to an evaluation platform, usually run by people hosting a shared task or leaderboard. In this setting, researchers submit their models or model outputs to the evaluation platform, which organizes and runs all the human evaluations. For example, GENIE \citep{Khashabi2021GENIEAL} and GEM \citep{Gehrmann2021TheGB} both include standardized human evaluations for understanding models' progress across several generation tasks.
ChatEval is an evaluation platform for open-domain chatbots based on both human and automatic metrics \citep{Sedoc2019ChatEvalAT}, and TuringAdvice \citep{Zellers2020EvaluatingMB} tests models' language understanding capabilities by having people read and rate the models' ability to generate advice.

Of course, as with all leaderboards and evaluation platforms, with uniformity and consistency come rigidity and the possibility of overfitting to the wrong objectives. Discussions of how to standardize human evaluations should take this into account. A person's goal when producing text can be nuanced and diverse, and the ways of evaluating text should reflect that.

\subsection{Extrinsic Evaluation}
\label{sec:extrinsic}
An \textit{extrinsic evaluation} measures how successful the system is in a downstream task. Extrinsic evaluations are the most meaningful evaluation as they show how a system actually performs in a downstream setting, but they can also be expensive and difficult to run \citep{Reiter2009AnII}. For this reason, intrinsic evaluations are more common than extrinsic evaluations \citep{gkatzia-mahamood-2015-snapshot, van-der-lee-etal-2019-best} and have become increasingly so, which \citet{van-der-lee-etal-2019-best} attribute to a recent shift in focus on NLG subtasks rather than full systems.

An NLG system's success can be measured from two different perspectives: a user's success in a task and the system's success in fulfilling its purpose \citep{Hastie2014ACE}. Extrinsic methods that measure a user's success at a task look at what the user is able to take away from the system, e.g., improved decision making or higher comprehension accuracy \citep{gkatzia-mahamood-2015-snapshot}. For example, \citet{Young1999UsingGM}, which \citet{Reiter2009AnII} point to as one of the first examples of extrinsic evaluation of generated text, evaluate automatically generated instructions by the number of mistakes subjects made when they followed them. 
System success-based extrinsic evaluations, on the other hand, measure an NLG system's ability to complete the task for which it has been designed. For example, \citet{Reiter2003LessonsFA} generate personalized smoking cessation letters and report how many recipients actually gave up smoking.
Post-editing, most often seen in machine translation \citep{Aziz2012PETAT,Denkowski2014LearningFP}, can also be used to measure a system's success by measuring how many changes a person makes to a machine-generated text.

Extrinsic human evaluations are commonly used in evaluating the performance of dialog systems \citep{surveydialogeval} and have made an impact on the development of the dialog modeling systems. Various approaches have been used to measure the system's performance when talking to people, such as measuring the conversation length or asking people to rate the system. The feedback is collected by real users of the dialog system \citep{sdc2010,label2000,zhou2020design} at the end of the conversation. 
The Alexa Prize\footnote{\url{https://developer.amazon.com/alexaprize}} follows a similar strategy by letting real users
interact with operational systems and gathering the user feedback over a span of several months. 
However, the most commonly used human evaluations of dialog systems is still via crowdsourcing platforms such as Amazon Mechanical Turk (AMT) \citep{mrnndialogres,peng2020few-shot,li2020multi-domain,zhou2020design}. %Such platforms provide human judges from different skill levels. 
\citet{humanuserseval2011} suggest that using enough crowdsourced users can yield a good quality metric, which is also comparable to the human evaluations in which subjects interact with the system and evaluate afterwards. 
%Most recent work mostly chose crowd-sourcing for human evaluation.

\subsection{The Evaluators}

For many NLG evaluation tasks, no specific expertise is required of the evaluators other than a proficiency in the language of the generated text. This is especially true when fluency-related aspects of the generated text are the focus of the evaluation. Often, the target audience of an NLG system is broad, e.g., a summarization system may want to generate text for anyone who is interested in reading news articles or a chatbot needs to carry out a conversation with anyone who could access it. In these cases, human evaluations benefit from being performed on as wide a population as possible. 

Evaluations can be performed either in-person or online. An in-person evaluation could simply be performed by the authors or a group of evaluators recruited by the researchers to come to the lab and participate in the study. The benefits of in-person evaluation are that it is easier to train and interact with participants, and that it is easier to get detailed feedback about the study and adapt it as needed. Researchers also have more certainty and control over who is participating in their study, which is especially important when trying to work with a more targeted set of evaluators. However, in-person studies can also be expensive and time-consuming to run. For these reasons, in-person evaluations tend to include fewer participants, and the set of people in proximity to the research group may not accurately reflect the full set of potential users of the system. In-person evaluations may also be more susceptible to response biases, adjusting their decisions to match what they believe to be the researchers' preferences or expectations \citep{Nichols2008TheGE, Orne1962OnTS}.

To mitigate some of the drawbacks of in-person studies, online evaluations of generated texts have become increasingly popular. While researchers could independently recruit participants online to work on their tasks, it is common to use crowdsourcing platforms that have their own users whom researchers can recruit to participate in their task, either by paying them a fee (e.g., Amazon Mechanical Turk\footnote{\url{https://www.mturk.com/}}) or rewarding them by some other means (e.g., LabintheWild\footnote{\url{http://www.labinthewild.org/}}, which provides participants with personalized feedback or information based on their task results). These platforms allow researchers to perform large-scale evaluations in a time-efficient manner, and they are usually less expensive (or even free) to run. They also allow researchers to reach a wider range of evaluators than they would be able to recruit in-person (e.g., more geographical diversity). However, maintaining quality control online can be an issue \citep{Ipeirotis2010QualityMO, oppenheimer2009instructional}, and the demographics of the evaluators may be heavily skewed depending on the user base of the platform \citep{Difallah2018DemographicsAD, Reinecke2015LabintheWildCL}. Furthermore, there may be a disconnect between what evaluators online being paid to complete a task would want out of an NLG system and what the people who would be using the end product would want. 

Not all NLG evaluation tasks can be performed by any subset of speakers of a given language. Some tasks may not transfer well to platforms like Amazon Mechanical Turk where workers are more accustomed to dealing with large batches of microtasks. Specialized groups of evaluators can be useful when testing a system for a particular set of users, as in extrinsic evaluation settings. 
Researchers can recruit people who would be potential users of the system, e.g., students for educational tools or doctors for bioNLP systems. Other cases that may require more specialized human evaluation are projects where evaluator expertise is important for the task or when the source texts or the generated texts consist of long documents or a collection of documents. Consider the task of citation generation \citep{Luu2020CitationTG}: given two scientific documents A and B, the task is to generate a sentence in document A that appropriately cites document B. To rate the generated citations, the evaluator must be able to read and understand two different scientific documents and have general expert knowledge about the style and conventions of academic writing. For these reasons, \citet{Luu2020CitationTG} choose to run human evaluations with expert annotators (in this case, NLP researchers) rather than crowdworkers.

\subsection{Inter-Evaluator Agreement}\footnote{Some of the other terms used are: \textit{inter-rater reliability}, \textit{inter-rater agreement}, \textit{inter-rater concordance}, \textit{inter-observer reliability}, etc. In text generation inter-evaluator or inter-rater agreement are the most commonly used terms.}
While evaluators\footnote{In text generation, `\textit{judges}' are also commonly used.} often undergo training to standardize their evaluations, evaluating generated natural language will always include some degree of subjectivity. Evaluators may disagree in their ratings, and the level of disagreement can be a useful measure to researchers. High levels of inter-evaluator agreement generally mean that the task is well-defined and the differences in the generated text are consistently noticeable to evaluators, while low agreement can indicate a poorly defined task or that there are not reliable differences in the generated text.

Nevertheless, measures of inter-evaluator agreement are not frequently included in NLG papers. Only 18\% of the 135 generation papers reviewed in \citet{Amidei2019AgreementIO} include agreement analysis (though on a positive note, it was more common in the most recent papers they studied). When agreement measures are included, agreement is usually low in generated text evaluation tasks, lower than what is typically considered ``acceptable'' on most agreement scales \citep{Amidei2018RethinkingTA, Amidei2019AgreementIO}. However, as \citet{Amidei2018RethinkingTA} point out, given the richness and variety of natural language, pushing for the highest possible inter-annotator agreement may not be the right choice when it comes to NLG evaluation.

While there are many ways to capture the agreement between annotators \citep{Banerjee1999BeyondKA}, we highlight the most common approaches used in NLG evaluation. For an in-depth look at annotator agreement measures in natural language processing, refer to \citet{Artstein2008InterCoderAF}.

\subsubsection{Percent agreement}
A simple way to measure agreement is to report the percent of cases in which the evaluators agree with each other. If you are evaluating a set of generated texts $X$ by having people assign a score to each text $x_i$, then let $a_i$ be the agreement in the scores for $x_i$ (where $a_i=1$ if the evaluators agree and $a_i=0$ if they don't). Then the percent agreement for the task is:
\begin{equation}
\label{eq:percentagree}
P_a = \frac{\sum_{i=0}^{|X|} a_i}{|X|}
\end{equation}

So $P_a=0$ means the evaluators did not agree on their scores for any generated text, while $P_a=1$ means they agreed on all of them.

However, while this is a common way people evaluate agreement in NLG evaluations \citep{Amidei2019AgreementIO}, it does not take into account the fact that the evaluators may agree purely by chance, particularly in cases where the number of scoring categories are low or some scoring categories are much more likely than others \citep{Artstein2008InterCoderAF}. We need a more complex agreement measure to capture this.
\subsubsection{Cohen's $\kappa$}
Cohen's $\kappa$ \citep{cohenskappa} is an agreement measure that can capture evaluator agreements that may happen by chance. In addition to $P_a$, we now consider $P_c$, the probability that the evaluators agree by chance. So, for example, if two evaluators ($e_1$ and $e_2$) are scoring texts $X$ with a score from the set $S$, then $P_c$ would be the odds of them both scoring a text the same:

\begin{equation}
\label{eq:percentchance}
P_c = \sum_{s \in S} P(s|e_1) * P(s|e_2)
\end{equation}

For Cohen's $\kappa$, $P(s|e_i)$ is estimated using the frequency with which the evaluator $e_i$ assigned each of the scores across the task.\footnote{There are other related agreement measures, e.g., Scott's $\pi$ \citep{scottspi}, that only differ from Cohen's $\kappa$ in how to estimate $P(s|e_i)$. These are well described in \citet{Artstein2008InterCoderAF}, but we do not discuss these here as they are not commonly used for NLG evaluations \citep{Amidei2019AgreementIO}.} 
%So, for example, if there are two scores, 0 and 1, and $e_1$ assigns 6 scores as 0s and 4 scores as 1s, and $e_2$ assigns 5 0s and 5 1s, then $P_c = 0.6*0.5 + 0.4*0.5$.
Once we have both $P_a$ and $P_c$, Cohen's $\kappa$ can then be calculated as:

\begin{equation} \label{eq:kappa}
    \kappa = \frac{P_a - P_c}{1 - P_c}
\end{equation}

\subsubsection{Fleiss' $\kappa$}
As seen in Equation \ref{eq:percentchance}, Cohen's $\kappa$ measures the agreement between two annotators, but often many evaluators have scored the generated texts, particularly in tasks that are run on crowdsourcing platforms. Fleiss' $\kappa$ \citep{fleissskappa} can measure agreement between multiple evaluators. This is done by still looking at how often pairs of evaluators agree, but now considering all possible pairs of evaluators. So now $a_i$, which we defined earlier to be the agreement in the scores for a generated text $x_i$, is calculated across all evaluator pairs:

\begin{equation}
    a_i = \frac{\sum_{s \in S} \text{\# of evaluator pairs who score $x_i$ as $s$}}{\text{total \# of evaluator pairs}}
\end{equation}

Then we can once again define $P_a$, the overall agreement probability, as it is defined in Equation \ref{eq:percentagree}---the average agreement across all the texts.

To calculate $P_c$, we estimate the probability of a judgment $P(s|e_i)$ by the frequency of the score across all annotators.
% and assuming each annotator is equally likely to draw randomly from this distribution. 
So if $r_s$ is the proportion of judgments that assigned a score $s$, then the likelihood of two annotators assigning score $s$ by chance is $r_s * r_s = r_s^2$. Then our overall probability of chance agreement is:

\begin{equation}
    P_c = \sum_{s \in S} r_s^2
\end{equation}

With these values for $P_a$ and $P_c$, we can use Equation \ref{eq:kappa} to calculate Fleiss' $\kappa$.

\subsubsection{Krippendorff's $\alpha$}
Each of the above measures treats all evaluator disagreements as equally bad, but in some cases, we may wish to penalize some disagreements more harshly than others. Krippendorff's $\alpha$ \citep{krippendorffsalpha}, which is technically a measure of evaluator \textit{disagreement} rather than agreement, allows different levels of disagreement to be taken into account.\footnote{Note that there are other measures that permit evaluator disagreements to be weighted differently. For example, weighted $\kappa$ \citep{weightedkappa} extends Cohen's $\kappa$ by adding weights to each possible pair of score assignments. In NLG evaluation, though, Krippendorff's $\alpha$ is the most common of these weighted measures; in the set of NLG papers surveyed in \citet{Amidei2019AgreementIO}, only one used weighted $\kappa$.}

Like the $\kappa$ measures above, we again use the frequency of evaluator agreements and the odds of them agreeing by chance. However, we will now state everything in terms of disagreement. First, we find the probability of disagreement across all the different possible score pairs $(s_m, s_n)$, which are weighted by whatever value $w_{m,n}$ we assign the pair. So: 

\begin{equation}
    P_d = \sum_{m=0}^{|S|} \sum_{n=0}^{|S|} w_{m,n} \sum_{i=0}^{|X|} \frac{\text{\# of evaluator pairs that assign } x_i \text{ as } (s_m, s_n)}{\text{total \# of evaluator pairs}}
\end{equation}

(Note that when $m==n$, i.e., the pair of annotators agree, $w_{m,n}$ should be 0.) 

Next, to calculate the expected disagreement, we make a similar assumption as in Fleiss' $\kappa$: the random likelihood of an evaluator assigning a score $s_i$ can be estimated from the overall frequency of $s_i$. If $r_{m,n}$ is the proportion of all evaluation pairs that assign scores $s_m$ and $s_n$, then we can treat it as the probability of two evaluators assigning scores $s_m$ and $s_n$ to a generated text at random. So $P_c$ is now:

\begin{equation}
    P_c = \sum_{m=0}^{|S|} \sum_{n=0}^{|S|} w_{m,n} r_{m,n}
\end{equation}

Finally, we can calculate Krippendorff's $\alpha$ as:

\begin{equation}
    \alpha = 1 - \frac{P_d}{P_c}
\end{equation}

\section{Untrained Automatic Evaluation Metrics}
\label{metric}

With the increase of the numbers of NLG applications 
and their benchmark datasets, the evaluation of NLG systems
% for NLG applications 
has become increasingly important. 
%Today, several metrics that focus on different task-specific or task-agnostic criteria are used to evaluate NLG system output, which is machine-generated text. 
Arguably, humans can evaluate most of the generated text with little effort\footnote{For some domains that require domain knowledge (e.g., factual correctness of a scientific article) or background knowledge (e.g., knowledge of a movie or recipe) might be necessary to evaluate the generated recipe obeys the actual instructions if the instructions are not provided}. However, human evaluation is costly and time-consuming to design and run, and more importantly, the results are not always repeatable \citep{belz-reiter-2006-comparing}. % \liz{empty ref}. 
Thus, automatic evaluation metrics are employed as an alternative in both developing new models and comparing them against state-of-the-art approaches. 
In this survey, we group automatic metrics into two categories: untrained automatic metrics that do not require training (this section), and machine-learned evaluation metrics that are based on machine-learned models 
% and require training of a model and a training dataset 
(Section~\ref{model}). 

Untrained automatic metrics for NLG evaluation are used to measure the effectiveness of the models that generate text, such as in machine translation, image captioning, or question generation.  These metrics compute a score that indicates the similarity (or dissimilarity) between an automatically generated text and human-written reference (gold standard) text. 
%Because untrained automatic evaluation metrics don't require training, they 
Untrained automatic evaluation metrics are fast and efficient and are widely used to quantify day-to-day progress of model development, e.g., comparing models trained with different hyperparameters. % and enable much faster experimentation. 
In this section we review untrained automatic metrics used in different NLG applications and briefly discuss the advantages and drawbacks of commonly used metrics.
We group the untrained automatic evaluation methods, as in Table~\ref{tab:autometrics}, into five categories: 
\begin{itemize}
    \setlength\itemsep{-0.5em}
    \item \textit{n}-gram overlap metrics
    \item distance-based metrics
    \item diversity metrics
    \item content overlap metrics
    \item grammatical feature based metrics\footnote{No specific metric is defined, mostly syntactic parsing methods are used as metric. See section~\ref{synsim} for more details.}
\end{itemize}
We cluster some of these metrics in terms of different efficiency criteria (where applicable) in Table~\ref{tab:summaryauto}\footnote{Recent work reports that with limited human studies most untrained automatic metrics have weaker correlation with human judgments and the correlation strengths would depend on the specific human's evaluation criteria \citep{shimorina2021human}. 
The information in Table~\ref{tab:summaryauto} relating to correlation with human judgments is  obtained from the published work which we discuss in this section. We suggest the reader refer to the model-based evaluation metrics in the next section, in which we survey evaluation models that have reported tighter correlation with the human judgments on some of the evaluation criteria.}. 
Throughout this section, we will provide a brief description of the selected untrained metrics as depicted in Table~\ref{tab:autometrics}, discuss about how they are used in evaluating different text generation tasks and provide references for others for further read. 
We will highlight some of their strengths and weaknesses in bolded sentences.

% ##########################################
\subsection{\textit{n}-gram Overlap Metrics for Content Selection}
% ##########################################
\textit{n}-gram overlap metrics are commonly used for evaluating  NLG systems and measure
the degree of ``matching'' between machine-generated and human-authored (ground-truth) texts. In this section we present several \textit{n}-gram match features and the NLG tasks they are used to evaluate.

\noindent\textbf{\textsc{f-score}.}
% % ################################
% \subsubsection{\textsc{f-score}}
% % ################################
Also called F-measure, the \textsc{f-score} is a measure of accuracy. It balances the generated text's precision and recall by the harmonic mean of the two measures. The most common instantiation of the \textsc{f-score} is the \textsc{f1-score} ($F_1$).
In text generation tasks such as machine translation or summarization, \textsc{f-score} gives an indication as to the quality of the generated sequence that a model will produce \citep{autoevalsmt2,fforsum}.
Specifically for machine translation, \textsc{f-score}-based metrics have been shown to be effective in evaluating translation quality. 

\begin{table*}[t]
    \centering
    \small
    \begin{tabular}{clllllllll}
        \multicolumn{1}{l}{}            & Metric & Property & MT & IC & SR & SUM & DG & QG & RG \\
        \hline
\multirow{15}{*}{\rotatebox[origin=c]{90}{\parbox[c]{3cm}{\centering \textit{n}-gram overlap}}} 
                                        & \textsc{f-score}  &   precision and recall   & \checkmark   & \checkmark   &  \checkmark  &  \checkmark   &  \checkmark  &  \checkmark & \checkmark \\
                                        & \textsc{bleu}    &  \textit{n}-gram precision        & \checkmark & \checkmark   &   &   & \checkmark   & \checkmark &  \checkmark  \\
                                        & \textsc{meteor}  &   \textit{n}-gram w/ synonym match       &  \checkmark  & \checkmark   &    &     &  \checkmark  & &   \\                                         
                                        % & \textsc{self-bleu}   &   \textit{n}-gram precision &  \checkmark  &    &    & \checkmark    &  \checkmark  &  &   \\
                                        & \textsc{cider}  &    \textit{tf-idf} weighted \textit{n}-gram sim.      &    & \checkmark    &    &     &    & &  \\ 
                                        & \textsc{nist}    &  \textit{n}-gram precision        &   \checkmark &    &    &     &    &    \\
                                        &  \textsc{gtm}    & \textit{n}-gram metrics         &  \checkmark &    &   &   &   &  &   \\
                                        & \textsc{hlepor}  &  unigrams harmonic mean         &   \checkmark &    &    &     &    &&   \\ 
                                        & \textsc{ribes}  &   unigrams harmonic mean        &    &    &    &     &    &&   \\ 
                                        & \textsc{masi}  &     attribute overlap         &    &    &    &     &    &&   \\ 
                                        & \textsc{wer}  &   \% of insert,delete, replace       &    &    &  \checkmark  &     &    & &   \\
                                        % & \textsc{med}   &     text based edit distance     & \checkmark   &    &    &     &    & &   \\
                                        & \textsc{ter}   &   translation edit rate       & \checkmark   &    &    &     &    & &   \\
                                        & \textsc{rouge}  &  \textit{n}-gram recall        &    &    &    & \checkmark    & \checkmark   & &   \\ 
                                        & \textsc{dice}  &   attribute overlap       &    &    &    &     &    &&   \\ 
                                    %   &  \textsc{ttr}    &          lexical diversity &  &    &   &   &   & \checkmark & \checkmark  \\
        \hline
\multirow{6}{*}{\rotatebox[origin=c]{90}{\parbox[c]{1cm}{\centering distance-based}}} & \textsc{edit dist.}    & cosine similarity          &   \checkmark  &  \checkmark   &   \checkmark  & \checkmark     &   \checkmark  & \checkmark    \\
                                        & \textsc{meant 2.0}    &  vector based similarity          &  \checkmark  &    &    &     &    & &   \\           
                                        & \textsc{yisi}   & weighted similarity           &    \checkmark &    &    &     &    & &   \\
                                        & \textsc{wmd}    &   EMD on words        &    & \checkmark   &    &  \checkmark   &    & &   \\
                                        & \textsc{smd}  &     EMD on sentences     &    &  \checkmark  &  \checkmark  & \checkmark    &    &   &\\
                                        & \textsc{Fréchet}    &    distributional similarity &    & \checkmark   &    &     &    & &   \\      
        \hline
\multirow{3}{*}{\rotatebox[origin=c]{90}{\parbox[c]{1cm}{\centering content overlap}}} & \textsc{pyramid}    &  content selection        &    &    &  \checkmark  &     &    &    \\
                                        & \textsc{spice}    &   scene graph similarity       &    & \checkmark   &    &     &    &   & \\
                                        & \textsc{spider}  &   scene graph similarity       &    & \checkmark   &    &     &    &  & \\
        \hline
\multirow{4}{*}{\rotatebox[origin=c]{90}{\parbox[c]{1cm}{\centering diversity}}} &    &          &    &    &    &     &    &    \\
                                        & \textsc{ttr}    &  richness of vocabulary       &    &    & \checkmark   &     &    &   & \\
                                        % & \textsc{spider}  &   scene graph similarity       &    & \checkmark   &    &     &    &  & \\
                                        & \textsc{self-bleu}    &  \textit{n}-gram precision        &  &    &   &   &   & \checkmark &   \\
                                        &    &  &    &   &    &     &    &   & \\

        \hline
\end{tabular}
    \caption{
    Untrained \textit{automatic} and \textit{retrieval-based} metrics based on string match, string distance, or context overlap. The acronyms for some of the NLP sub-research fields that each metric is commonly used to evaluate text generation are: \textbf{MT}: Machine Translation, \textbf{IC}: Image Captioning, \textbf{SR}: Speech Recognition, \textbf{SUM}: Summarization, \textbf{DG}: Document or Story Generation, Visual-Story Generation, \textbf{QG}: Question Generation, \textbf{RG}: Dialog Response Generation. EMD:Earth Mover's Distance; Sim.: Similarity.
    }
    \label{tab:autometrics}
\end{table*}

\noindent\textbf{\textsc{bleu}.}
% ################################
% \subsubsection{\textsc{bleu}}
% ################################
The Bilingual Evaluation Understudy (\textsc{bleu}) is one of the first metrics used to measure the similarity between two sentences \citep{bleumetric}. Originally proposed for machine translation, it compares a candidate translation of text to one or more reference translations. \textsc{bleu} is a weighted geometric mean of \textit{n}-gram precision scores.

It has been argued that although \textsc{bleu} has significant advantages, it may not be the ultimate measure for improved machine translation quality \citep{Callison-burch06re-evaluatingthe}.
While earlier work has reported that \textsc{bleu} correlates well with human judgments \citep{Lee2005NIST2M,denoual-lepage-2005-bleu}, more recent work argues that although it can be a good metric for the machine translation task \citep{Zhang04interpretingbleunist} for which it is designed, it doesn't correlate well with human judgments for other generation tasks (such as image captioning or dialog response generation). \citet{reiter-2018-structured} reports that there’s not enough evidence to support that \textsc{bleu} is the best metric for evaluating NLG systems other than machine translation.
\citet{langaugegan} found that generated text with perfect \textsc{bleu} scores was often grammatically correct but lacked semantic or global coherence, concluding that the generated text has poor information content.

Outside of machine translation, \textsc{bleu} has been used for other text generation tasks, such as document summarization \citep{graham-2015-evaluating}, image captioning \citep{showandtell}, 
human-machine conversation \citep{Gao2018},
and language generation \citep{semeniuta2019on}.
In \cite{graham-2015-evaluating}, it was concluded that \textsc{bleu} 
achieves strongest correlation with human assessment, but does not significantly outperform
the best-performing \textsc{rouge} variant. 
A more recent study has demonstrated that \textit{n}-gram matching scores such as \textsc{bleu} can be an
insufficient and potentially less accurate metric for unsupervised language generation \citep{semeniuta2019on}.

Text generation research, especially when focused on short text generation like sentence-based machine translation or question generation, has successfully used \textsc{bleu} for benchmark analysis with models since it is fast, easy to calculate, and enables a comparison with other models on the same task. 
However, \textsc{bleu} has some drawbacks for NLG tasks where contextual understanding and reasoning is the key (e.g., story generation \citep{storygen1,storygen2} or long-form question answering \citep{ELI5}). It considers neither semantic meaning nor %directly consider 
sentence structure. It does not handle morphologically rich languages well, nor does it map well to human judgments \citep{rachelblog}. Recent work by \citet{mathur2020tangled} investigated how sensitive the machine translation evaluation metrics are to outliers. 
They found that when there are outliers in tasks like machine translation, metrics like \textsc{bleu} lead to high correlations yielding false conclusions
about reliability of these metrics. They report that when the outliers are removed, these metrics do not correlate as well as before, which adds evidence to the unreliablity of \textsc{bleu}. 

We will present other metrics that address some of these shortcomings throughout this paper.

\noindent\textbf{\textsc{rouge}.}
% \subsubsection{\textsc{rouge}}
\label{rougescore}
Recall-Oriented Understudy for Gisting Evaluation (\textsc{rouge}) \citep{lin-2004-rouge} is a set of metrics for evaluating automatic summarization of long texts consisting of multiple sentences or paragraphs. Although mainly designed for evaluating single- or multi-document summarization, it has also been used for evaluating short text generation, such as machine translation \citep{lin-och-2004-automatic}, image captioning \citep{imageeval}, and question generation \citep{qge,unilm}.
\textsc{rouge} includes a large number of distinct variants, including eight different \textit{n}-gram counting
methods to measure \textit{n}-gram overlap between the generated and the ground-truth (human-written) text: \textbf{\textsc{rouge-\{1/2/3/4\}}} measures the overlap of \textit{unigrams}/\textit{bigrams}/\textit{trigrams}/\textit{four-grams} (single tokens) between the reference and hypothesis text (e.g., summaries); \textbf{\textsc{rouge-l}} measures the longest matching sequence of words using longest common sub-sequence (LCS); \textbf{\textsc{rouge-s}} (less commonly used) measures skip-bigram\footnote{A \textit{skip-gram} \citep{Huang92thesphinx-ii} is a type of \textit{n}-gram in which tokens (e.g., words) don't need to be consecutive but in order in the sentence, where there can be gaps between the tokens that are skipped over. In NLP research, they are used to overcome data sparsity issues.}-based co-occurrence statistics; \textbf{\textsc{rouge-su}} (less commonly used) measures skip-bigram and unigram-based co-occurrence statistics. 

Compared to \textsc{bleu}, \textsc{rouge} focuses on recall rather than precision and is more interpretable than \textsc{bleu} \citep{Callison-burch06re-evaluatingthe}.
Additionally, \textsc{rouge} includes the mean or median score from individual output text, which allows for a significance test of differences in system-level \textsc{rouge} scores, while this is restricted in 
\textsc{bleu} \citep{graham-baldwin-2014-testing,graham-2015-evaluating}.
%\textsc{rouge} evaluates the adequacy of the generated output text by counting how many \textit{n}-grams in the generated output text match the \textit{n}-grams in the reference (human-generated) output text.
However, \textsc{rouge}'s reliance on \textit{n}-gram matching can be an issue, especially for long-text generation tasks \citep{kilickaya_2016}, because it doesn't provide information about the narrative flow, grammar, or topical flow of the generated text, nor does it evaluate the factual correctness of the text compared to the corpus it is generated from.

\noindent\textbf{\textsc{meteor}.}
% \subsubsection{\textsc{meteor}}
The Metric for Evaluation of Translation with Explicit ORdering (\textsc{meteor}) \citep{Lavie2004TheSO,banerjee-lavie-2005-meteor} is a metric designed to address some of the issues found in \textsc{bleu} and has been widely used for evaluating machine translation models and other models, such as image captioning \citep{kilickaya_2016}, question generation \citep{qge,qglearningtoask}, and summarization \citep{see-etal-2017-get,fastabstsumm,Yan2020ProphetNetPF}.
Compared to \textsc{bleu}, which only measures precision, \textsc{meteor} is based on the harmonic mean of the unigram precision and recall, in which recall is weighted higher than precision. Several metrics support this property since it yields high correlation with human judgments \citep{meteorcorr}.

\textsc{meteor} has several variants that extend exact word matching that most of the metrics in this category do not include, such as stemming and synonym matching. These variants address the problem of reference translation variability, allowing for morphological variants and synonyms to be recognized as valid translations. The metric has been found to produce good correlation with human judgments at the sentence or segment level \citep{meteoreval}. This differs from \textsc{bleu} in that \textsc{meteor} is explicitly designed to compare at the sentence level rather than the corpus level.

\noindent\textbf{\textsc{cide}r.}
Consensus-based Image Description Evaluation (\textsc{cide}r) is an automatic metric for measuring the similarity of a generated sentence against a set of human-written sentences using a consensus-based protocol.
Originally proposed for image captioning \citep{cider}, \textsc{cide}r shows high agreement with consensus as assessed by humans. It enables a comparison of text generation models based on their ``human-likeness,'' without having to create arbitrary calls on weighing content, grammar, saliency, etc. with respect to each other.

The \textsc{cide}r metric presents three explanations about what a hypothesis sentence should contain: (1) \textit{n}-grams in the hypothesis sentence should also occur in the reference sentences, (2) If an \textit{n}-gram does not occur in a reference sentence, it shouldn't be in the hypothesis sentence, (3) \textit{n}-grams that commonly occur across all image-caption pairs in the dataset should be assigned lower weights, since they are potentially less informative. While \textsc{cide}r has been adopted as an evaluation metric for image captioning and has been shown to correlate well with human judgments on some datasets (PASCAL-50S and ABSTRACT-50S datasets) \citep{cider}, recent studies have shown that metrics that include semantic content matching such as \textsc{spice} can correlate better with human judgments \citep{spice, spider}. %We will review these 

\noindent\textbf{\textsc{nist}}. 
Proposed by the US National Institute of Standards and Technology, \textsc{nist} \citep{Martin00thenist} is a method similar to \textsc{bleu} for evaluating the quality of text. Unlike \textsc{bleu}, which treats each \textit{n}-gram equally, \textsc{nist} heavily weights \textit{n}-grams that occur less frequently, as co-occurrences of these \textit{n}-grams are more informative than common \textit{n}-grams \citep{autoevalsmt}. 

\noindent\textbf{\textsc{gtm}}. 
The \textsc{gtm} metric \citep{Turian2003} measures \textit{n}-gram similarity between the model-generated hypothesis translation and the reference sentence by using precision, recall and F-score measures.

\noindent\textbf{\textsc{hlepor}}. 
Harmonic mean of enhanced Length Penalty, Precision, \textit{n}-gram Position
difference Penalty, and Recall (\textsc{hlepor}), initially proposed for machine translation, is a metric designed for morphologically complex languages like Turkish or Czech \citep{hlepor}.
Among other factors, \textsc{hlepor} uses part-of-speech tags' similarity to capture syntactic information.

\noindent\textbf{\textsc{ribes}}. 
Rank-based Intuitive Bilingual Evaluation Score (\textsc{ribes}) \citep{isozaki-etal-2010-automatic} is another untrained automatic evaluation metric for machine translation. It was developed by NTT Communication Science Labs  and designed to be more informative for Asian languages--―like Japanese and Chinese---since it doesn’t rely on word boundaries. Specifically, \textsc{ribes} is based on how the words in generated text are ordered. It uses the rank correlation coefficients measured based on the word order from the hypothesis (model-generated) translation and the reference translation. 

\noindent\textbf{\textsc{dice} and \textsc{masi}}. 
Used mainly for referring expression generation evaluation, \textsc{dice} \citep{gatt-etal-2008-tuna} measures the overlap of a set of attributes between the human-provided referring expression and the model-generated expressions. The expressions are based on an input image (e.g., the large chair), and the attributes are extracted from the expressions, such as the \textit{type} or \textit{color} \citep{chen-van-deemter-2020-lessons}. The \textsc{masi} metric \citep{gatt-etal-2008-tuna}, on the other hand, adapts the Jaccard coefficient, which biases it in favour of similarity when a set of attributes is a subset of the other attribute set. 

\begin{table*}[t]
    \centering
    \small
    \begin{tabular}{p{0.40\linewidth} | p{0.55\linewidth}}
        Evaluation Criteria & Evaluation Metric \\
        \hline
        measures semantic similarity (content) & \textsc{pyramid}, \textsc{spice}, \textsc{spider}, \textsc{yisi}, \textsc{sps}, \textsc{te} \\
        % measures narrative coherence and cohesion & \\
        measures diversity & \textsc{wmd}, \textsc{smd}, \textsc{ttr}, \textsc{self-bleu} \\
        measures fluency & \textsc{bleu}, \textsc{rouge}, \textsc{nist}\\
        punishes length differences & \textsc{f-score}, \textsc{bleu}, \textsc{nist}, \textsc{rouge} \\
        punishes grammatical errors & \textsc{meteor}, \textsc{nist}  \\
        correlates well with human judgments & \textsc{meteor}, \textsc{spice}, \textsc{ter}  \\
        \hline
\end{tabular}
    \caption{Clustering several of the untrained metrics based on different criteria.
    }
    \label{tab:summaryauto}
\end{table*}

% ##########################################
\subsection{Distance-Based Evaluation Metrics for Content Selection}
% ##########################################
A distance-based metric in NLG applications uses a distance function to measure the similarity between two text units %relationship between discrete tokens in a sequence 
(e.g., words, sentences).
First, we represent two text units using vectors. 
Then, we compute the distance between the vectors. 
The smaller the distance, the more similar the two text units are.
This section reviews distance-based similarity measures
% used to evaluate NLG models, 
where text vectors can be constructed using discrete tokens, such as bag of words (\S\ref{sec:discrete-dist}) or embedding vectors (\S\ref{sec:embedding-dist}). 
We note that even though the embeddings that are used by these metrics to represent the text vectors are pre-trained, these metrics are not trained to mimic the human judgments, as in the machine-learned metrics that we summarize in Section \ref{model}. 

%##################################
\subsubsection{Edit Distance-Based Metrics}
\label{sec:discrete-dist}
%##################################
Edit distance, one of the most commonly used evaluation metrics in natural language processing, measures how dissimilar two text units are based on the minimum number of operations required to transform one text into the other. We summarize some of the well-known edit distance measures below.

% ###################################
\paragraph{\textsc{wer}}
% ###################################
Word error rate (\textsc{wer}) has been commonly used for measuring the performance of speech recognition systems, as well as to evaluate the quality of machine translation systems \citep{tomas-etal-2003-quantitative}. 
%This metric measures the number of words that differ between a piece of machine-translated text and a reference translation. 
Specifically, \textsc{wer} is the percentage of words that need to be inserted, deleted, or replaced in the translated sentence to obtain the reference sentence, i.e., the edit distance between the reference and hypothesis sentences. 

\textsc{wer} has some limitations. 
For instance, while its value is lower-bounded by zero, which indicates a perfect match between the hypothesis and reference text, its value is not upper-bounded, making it hard to evaluate in an absolute manner \citep{werissue}. 
It is also reported to suffer from weak correlation with human evaluation. 
For example, in the task of spoken document retrieval, 
% which is a voice search task, it was reported that 
the \textsc{wer} of an automatic speech recognition system is reported to poorly correlate with the retrieval system performance \citep{asrwererrors}.

% ###################################
\paragraph{\textsc{ter}}
% ###################################
Translation edit rate (\textsc{ter}) \citep{Snover06astudy} is defined as the minimum number of edits
needed to change a generated text so that it exactly
matches one of the references, normalized by the
average length of the references. 
While \textsc{ter} has been shown to correlate well with human judgments in evaluating machine translation quality, it suffers from some limitations. %has several flaws. 
For example, it can only capture similarity in a narrow sense, as it only uses a single reference translation and considers only exact word matches between the hypothesis and the reference. 
This issue can be partly addressed by constructing a lattice of reference translations, a technique that has been used to combine the output of multiple translation systems \citep{Rosti_improvedword-level}.

% ########################################
\subsubsection{Vector Similarity-Based Evaluation Metrics}
\label{sec:embedding-dist}

In NLP, embedding-based similarity measures are commonly used in addition to \textit{n}-gram-based similarity metrics. Embeddings are real-valued vector representations of character or lexical units, such as word-tokens or \textit{n}-grams, that allow tokens with similar meanings to have similar representations. Even though the embedding vectors are learned using supervised or unsupervised neural network models, the vector-similarity metrics we summarize below assume the embeddings are pre-trained and simply used as input to calculate the metric.  

% ##################################
\paragraph{\textsc{meant 2.0}}
%###################################
The vector-based similarity measure \textsc{meant} uses word embeddings and shallow semantic parses to compute lexical and structural similarity \citep{Lo2017MEANT2A}. It evaluates translation adequacy
by measuring the similarity of the semantic frames and their role fillers between the human references
and the machine translations.

%##################################
\paragraph{\textsc{yisi}} 
%##################################
Inspired by the \textsc{meant} score, \textsc{yisi}\footnote{YiSi, is the romanization of the Cantonese word 
\begin{CJK*}{UTF8}{gbsn}
意思
\end{CJK*}, which translates as `meaning' in English.} \citep{lo-2019-yisi} is proposed to evaluate the accuracy of machine translation model outputs. It is based on the weighted distributional lexical semantic similarity, as well as shallow semantic structures. Specifically, it extracts the longest common character sub-string from the hypothesis and reference translations to measure the lexical similarity. 

% ##################################
\paragraph{Word Mover's Distance (\textsc{wmd})}
%###################################
Earth mover's distance (\textsc{emd}), also known as the Wasserstein metric \citep{EarthMoversDistanceRubner98}, is a measure of the distance between two probability distributions. 
Word mover's distance (\textsc{wmd}; \citealp{kusner2015doc}) is a discrete version of \textsc{emd} that calculates the distance between two sequences (e.g., sentences, paragraphs, etc.), each represented with relative word frequencies. It combines item similarity\footnote{The similarity can be defined as cosine, Jaccard, Euclidean, etc.} on bag-of-word (BOW) histogram representations of text \citep{Goldberg2018}
with word embedding similarity. In short, 
\textsc{wmd} has several intriguing properties:
\begin{itemize}
    \setlength\itemsep{-0.5em}
    \item It is hyperparameter-free and easy to use.
    \item It is highly interpretable as the distance between two documents can be broken down and explained as the sparse distances between few individual words.
    \item It uses the knowledge encoded within the word embedding space, which leads to high retrieval accuracy.
\end{itemize}

\begin{figure*}[ht!]
    \centering
    \includegraphics[width=0.9\textwidth]{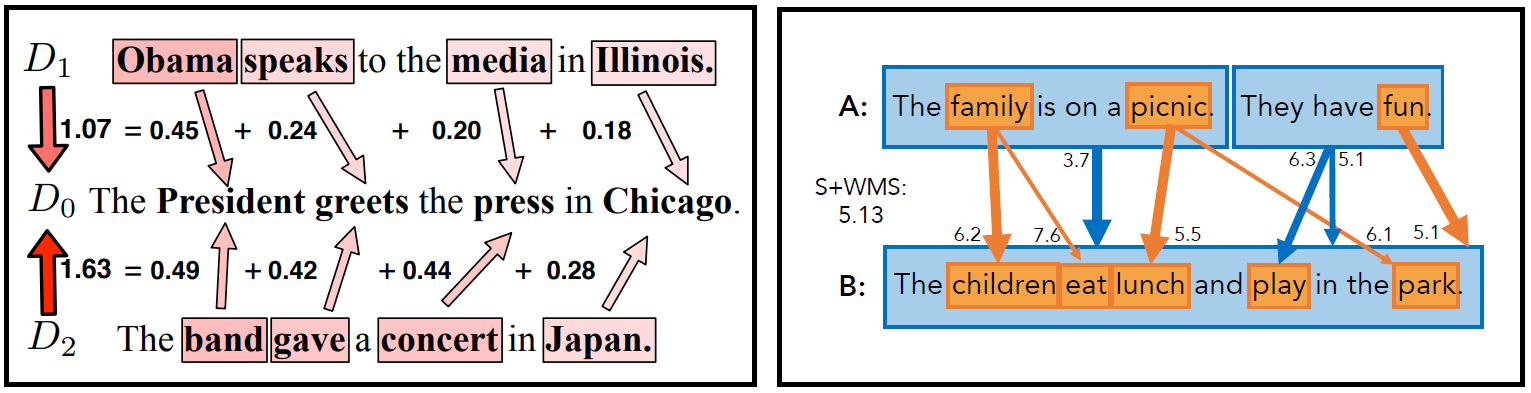}
    \caption{(LEFT) Illustration of Word Mover's Distance (WMD). Picture source: \citep{kusner2015doc}; (RIGHT) Illustration of Sentence Mover's Distance (SMD). Picture source: \citep{Clark2019SentenceMS}.}
    \label{fig:wmd}
\end{figure*}

Empirically, \textsc{wmd} has been instrumental to the improvement of many NLG tasks, specifically sentence-level tasks, such as image caption generation \citep{kilickaya_2016} and natural language inference \citep{EntailmentSulea2017}. 
However, while \textsc{wmd} works well for short texts, 
its cost grows prohibitively as the length of the documents increases, and the BOW approach can be problematic when documents become large as the relation between sentences is lost. 
By only measuring word distances, the metric cannot capture information conveyed in the group of words, for which we need higher-level document representations \citep{docembedding2015}.

%##################################
\paragraph{Sentence Mover's Distance (\textsc{smd})}
%##################################
Sentence Mover’s Distance (\textsc{smd}) is an automatic metric based on \textsc{wmd} %which is proposed as an extension of \textsc{wmd} 
to evaluate text in a continuous space using sentence embeddings \citep{Clark2019SentenceMS,Zhao2019}. 
%Coincidentally, the same year a similar metric with the same name has also been introduced in \citep{Zhao2019}.
%\textsc{smd} has been used to compare the generated texts to reference texts in tasks like machine translation and summarization, and is found to be correlated with human evaluation.
\textsc{smd} represents each document as a collection of sentences or of both words and  sentences (as  seen  in  Figure~\ref{fig:wmd}), where each sentence embedding is weighted according to its length. 
\textsc{smd} measures the cumulative distance of moving the sentences embeddings in one document to match another document's sentence embeddings. 
On a summarization task, \textsc{smd} correlated better with human judgments than \textsc{rouge} \citep{Clark2019SentenceMS}. 

\citet{Zhao2019} proposed a new version of \textsc{smd} %is investigated a metric extending \textsc{wmd} 
that attains higher correlation with human judgments. 
Similar to \textsc{smd}, they used word and sentence embeddings by taking the average of the token-based embeddings before the mover's distance is calculated. 
They also investigated different contextual embeddings models including ELMO and BERT by taking the power mean (which is an embedding aggregation method) of their embeddings at each layer of the encoding model.

% ##########################################
\subsection{\textit{n}-gram-Based Diversity Metrics}
% ##########################################

The lexical diversity score measures the breadth and variety of the word usage in writing \citep{textinspector}. 
% Consider two pieces of texts about in-class teaching. 
% The first repeatedly uses the same words such as `\textit{teacher}', `\textit{reads}', and `\textit{asks}'. 
% The second one avoids repetition by using different words or expressions, e.g, `\textit{lecturer}', `\textit{instructor}', `\textit{delivers}', `\textit{teaches}', `\textit{questions}', `\textit{explains}', etc. 
% The second text is more lexically diverse, 
Lexical diversity is desirable in many NLG tasks, such as 
conversational bots \citep{li-etal-2018-generating-reasonable},
story generation \citep{plotmachines}, 
question generation \citep{qglearningtoask,recentqg}, and abstractive question answering (Fan et al., 2019). 
%, indicating it is a more complex and challenging text.
Nevertheless, diversity-based metrics are rarely used on their own, as text diversity can come at the cost of text quality \citep{DBLP:journals/corr/abs-1904-03971, hashimoto-etal-2019-unifying,zhang-etal-2021-trading}, and some NLG tasks do not require highly diverse generations. For example, \citet{REITER2005137} reported that a weather forecast system was preferred over human meteorologists as the system produced report has a more consistent use of certain classes of expressions relating to reporting weather forecast.

In this section we review some of the metrics designed to measure the quality of the generated text in terms of lexical diversity. 

% #################################
\paragraph{Type-Token Ratio (\textsc{ttr})} 
% #################################
is a measure of lexical diversity \citep{ttr}, mostly used in linguistics to determine the richness of a writer’s or speaker’s vocabulary. It is computed as the number of unique words (types) divided by the total number of words (tokens) in a given segment of language.

Although intuitive and easy to use, \textsc{ttr} is sensitive to text length because the longer the document, the lower the likelihood that a token will be a new type. This causes the \textsc{ttr} to drop as more words are added.
To remedy this, a diversity metric, \textsc{hd-d} (hyper-geometric distribution function), was proposed to compare texts of different lengths \citep{MTLD}.

Measuring diversity using \textit{n}-gram repetitions is a more generalized version of \textsc{ttr}, which has been use for text generation evaluation.  
\citet{li-etal-2016-diversity} has shown that modeling mutual information between source and targets significantly decreases the chance of generating bland responses and improves the diversity of responses.
They use \textsc{bleu} and distinct word unigram and bigram counts to evaluate the proposed diversity-promoting objective function for dialog response generation.

% ################################
\paragraph{\textsc{Self-bleu}}
% ################################
\citet{texygen} proposed \textsc{self-bleu} as a diversity evaluation metric, which
calculates a \textsc{bleu} score for every generated sentence, treating the other generated sentences as references.
The average of these \textsc{bleu} scores is the \textsc{self-bleu} score of the text, where a lower \textsc{self-bleu} score implies higher diversity.
Several NLG papers have reported that \textsc{self-bleu} achieves good generation diversity \citep{texygen, NIPS2018_7717,neuraltextgen}. 
However, others have reported some weakness of the metric in generating diverse output \citep{langaugegan} or detecting mode collapse \citep{semeniuta2019on} in text generation with GAN \citep{goodfellow2014generative} models.

% ##########################################
\subsection{Explicit Semantic Content Match Metrics}
% ##########################################
Semantic content matching metrics define the similarity between human-written and model-generated text by extracting explicit semantic information units from text beyond \textit{n}-grams. These metrics operate on semantic and conceptual levels and are shown to correlate well with human judgments. We summarize some of them below.

\paragraph{\textsc{pyramid}}
% \subsubsection{\textsc{pyramid}}
% ########################################
The \textsc{pyramid} method is a semi-automatic evaluation method \citep{Nenkova2004} for evaluating the performance of document summarization models. 
Like other untrained automatic metrics that require references, this untrained metric also requires human annotations. 
It identifies summarization content units (SCUs) to compare information in a human-generated reference summary to the model-generated summary. 
To create a pyramid, annotators
%begin with model-generated summaries of the same source texts and 
select sets of text spans that express the same meaning across summaries, and
%Each set is referred to as a SCU and receives a label for mnemonic purposes. 
each SCU is a weighted according to the number of summaries that express the SCU’s meaning.

The \textsc{pyramid} metric relies on manual human labeling effort, which makes it difficult to automate.
%In a recent study, the 
\textsc{peak}: Pyramid Evaluation via Automated Knowledge Extraction \citep{peak} was presented as a fully automated variant of \textsc{pyramid} model, which can automatically assign the pyramid weights and was shown to correlate well with human judgments. 

\paragraph{\textsc{spice}}

\begin{figure*}
    \centering
    \includegraphics[width=0.8\textwidth]{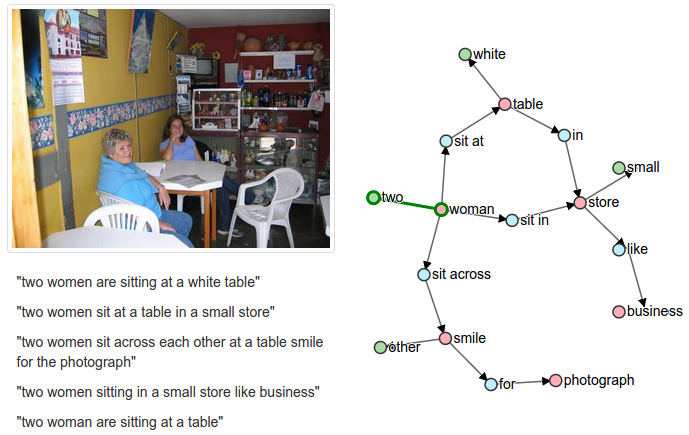}
    \caption{Illustration of Scene Graph Extraction for measuring the \textsc{spice} metric. A scene graph (right) is parsed from a set of reference image
captions on the left. Picture source: \citep{spice}.}
    \label{fig:sg}
\end{figure*}

Semantic propositional image caption evaluation (\textsc{spice}) \citep{spice} is an image captioning metric that measures the similarity between a list of reference human written captions 
$S = \left\{s_1,\cdots,s_m\right\}$ of an image and a hypothesis caption $c$ generated by a model.
%Instead of directly comparing a generated caption to a set of references in terms of syntactic agreement, \textsc{spice} parses each reference to derive an abstract scene graph representation. The generated caption is also parsed and compared to the scene graph to capture the semantic similarity. 
Instead of directly comparing the captions' text, \textsc{spice} parses each caption to derive an abstract scene graph representation, encoding the objects, attributes, and relationships detected in image captions \citep{schuster-etal-2015-generating}, as shown in Figure~\ref{fig:sg}. \textsc{spice} then computes the \textsc{f-score} using the hypothesis and reference scene graphs over the conjunction of logical tuples representing semantic propositions in the scene graph to measure their similarity.
%In \cite{spice}, the 
\textsc{spice} has been shown to have a strong correlation with human ratings.

Compared to \textit{n}-gram matching methods, \textsc{spice} 
can capture a broader sense of semantic similarity between a hypothesis and a reference text by using scene graphs. % to capture the semantic similarity. 
However, even though \textsc{spice} correlates well with human evaluations,
a major drawback is that it ignores the fluency of the generated captions \citep{sharif-etal-2018-learning}.

% ########################################
\paragraph{\textsc{spider}}
% #########################################
\citet{spider} proposed \textsc{spider}, which is a
linear combination of \textsc{spice} and \textsc{cider}. 
They show that optimizing \textsc{spice} alone often results in captions that are wordy and repetitive 
%They argue that this was due to 
because while scene graph similarity is good at measuring the semantic similarity between captions, it does not take into account the syntactical aspects of texts. 
Thus, a combination of semantic graph similarity (like \textsc{spice}) and \textit{n}-gram similarity measure (like \textsc{cider}) yields a more complete quality evaluation metric. 
However, the correlation of \textsc{spider} and human evaluation is not reported.
%was not evaluated for its with human judgments. 

% ########################################
\subsubsection{Semantic Similarity Models used as Evaluation Metrics}
% ########################################
Other text generation work has used the confidence scores obtained from semantic similarity methods as an evaluation metric. Such models can evaluate a reference and a hypothesis text based on their task-level semantics. The most commonly used methods based on the sentence-level similarity are as follows:

\begin{itemize}
    \item \textbf{Semantic Textual Similarity} (STS) is concerned with the degree of equivalence in the underlying semantics of paired text \citep{Agirre_semneval2016}. STS is used as an evaluation metric in text generation tasks such as machine translation, summarization, and dialogue response generation in conversational systems. The official score is based on weighted Pearson correlation between predicted similarity and human-annotated similarity. The higher the score, the better the the similarity prediction result from the algorithm \citep{maharjan-etal-2017-dt, semeval2017_cer}.  
    \item \textbf{Paraphrase identification} (PI) considers if two sentences express the same meaning \citep{dolan2005automatically,Barzilay+Lee:03a}. PI is used as a text generation evaluation score based on the textual similarity \citep{Kauchak2006} of a reference and hypothesis by finding a paraphrase of the reference sentence that is closer in wording to the hypothesis output. For instance, given the pair of sentences:
    
    \begin{quote}
\small
	reference: ``\textit{However, Israel’s reply failed to completely clear the U.S. suspicions.}''\\
	hypothesis: ``\textit{However, Israeli answer unable to fully remove the doubts.}''
	\end{quote}
    
    PI is concerned with learning to transform the reference sentence into:
    
     \begin{quote}
\small
	paraphrase: ``\textit{However, Israel’s \textit{answer} failed to completely \textit{remove} the U.S. suspicions.}''
		\end{quote}
    
    which is closer in wording to the hypothesis. In \citet{jiang2019tiger}, a new paraphrasing evaluation metric, \textsc{tiger}, is used for image caption generation evaluation.
    % (See details in section~\ref{nontext2textgen}.\liz{broken ref}) 
    Similarly, \cite{Lixin2019} introduce different strategies to select useful visual paraphrase pairs for training by designing a variety of scoring functions.
    \item \textbf{Textual entailment} (TE) is concerned with whether a hypothesis can be inferred from a premise, requiring understanding of the semantic similarity between the hypothesis and the premise (Dagan et al., 2006; Bowman et al., 2015). It has been used to evaluate several text generation tasks, including machine translation \citep{temt}, document summarization \citep{tesumm}, language modeling \citep{liu2019multi-task}, and video captioning \citep{pasunuru2017}. 
    \item \textbf{Machine Comprehension} (MC) is concerned with the sentence matching between a passage and a question, pointing out the text region that contains the answer \citep{SQuAD}. MC has been used for tasks like improving question generation \citep{yuan-etal-2017-machine,qglearningtoask} and document summarization \citep{teachingmachines}. 

\end{itemize}

% ##########################################
\subsection{Syntactic Similarity-Based Metrics}
\label{synsim}
% ##########################################

% \subsubsection{Syntactic Similarity}
A syntactic similarity metric captures the similarity between a reference and a hypothesis text at a structural level to capture the overall grammatical or sentence structure similarity. % some of which we summarize here.  

In corpus linguistics, part of speech (POS) tagging is the process of assigning a part-of-speech tag (e.g., verb, noun, adjective, adverb, and preposition, etc.) to each word in a sentence, based on its context, morphological behaviour, and syntax. POS tags have been commonly used in machine translation evaluation to evaluate the quality of the generated translations. \textsc{tesla} \citep{Dahlmeier_teslaat} was introduced as an evaluation metric to combine the synonyms of bilingual phrase tables and POS tags, while others use POS \textit{n}-grams together with a combination of morphemes and lexicon probabilities to compare the target and source translations \citep{Popovic2011,han2013}. 
POS tag information has been used for other text generation tasks such as story generation \citep{agirrezabal-etal-2013-pos}, summarization \citep{possumm}, and question generation \citep{zerrposqg}.

%\textbf{Sentence structure} through 
Syntactic analysis studies the arrangement of words and phrases in well-formed sentences. 
For example, a dependency parser extracts a dependency tree of a sentence to represent its grammatical structure.
%, as well as defines the relationships between “head” words and dependent words, which modify those heads. 
Several text generation tasks have enriched their evaluation criteria by leveraging syntactic analysis. 
%sentence structure information and dependency parsing. 
In machine translation, \citet{liu-gildea-2005-syntactic} used constituent labels and head-modifier dependencies to extract structural information from sentences for evaluation, while others use shallow parsers \citep{Lo2012FullyAS} or dependency parsers \citep{yu-etal-2014-red,amtdp}. 
\citet{yoshida-etal-2014-dependency} combined a sequential decoder with a tree-based decoder in a neural architecture for abstractive text summarization. 
%to generate the summary sentences and their syntactic parse. 

\section{Machine-Learned Evaluation Metrics}
\label{model}

%As we have summarized in Chapter \ref{metric}, automatic evaluation metrics have been widely used to evaluate NLG tasks. 
% Most of these metrics serve for several generation tasks efficiently. 
%For example, \textsc{bleu} \citep{bleumetric} and \textsc{meteor} \citep{meteoreval} are now standard metrics for evaluating machine translation models. \textsc{rouge} \citep{lin-2004-rouge} is mostly used for automatic summarization task evaluation. These metrics are also adapted for other multi-modal tasks such as video summarization \citep{videosumm}, image captioning \citep{imageeval}, etc. 
% Although useful, the main drawback of 
Many of the untrained evaluation metrics described in Section \ref{metric} assume that the generated text 
%is available, has acceptably good quality (e.g., coherent, non-repeating, has natural flow of events, objects and actions, etc.), and 
has significant word (or \textit{n}-gram) overlap with the ground-truth text. 
However, this assumption does not hold for NLG tasks that permit significant diversity and allow multiple plausible outputs for a given input (e.g.,  a social chatbot). 
%This is mainly observed in almost all text generation tasks such as image captioning, dialog response generation, and summarization. 
Table~\ref{table:evalerrors} shows two examples from the dialog response generation and image captioning tasks, respectively. 
In both tasks, the model-generated outputs are plausible given the input, but they do not share any words with the ground-truth output. % or have the same semantic meaning. 

One solution to this problem is to use embedding-based metrics, which measure semantic similarity rather than word overlap, as in Section \ref{sec:embedding-dist}. 
But embedding-based methods cannot help in situations when the generated output is semantically different from the reference, as in the dialog example. 
In these cases,
% researchers can use a machine-learned model to measure the semantic similarity between the system output and the reference. 
% In fact, 
we can build machine-learned models (trained on human judgment data) to mimic human judges to measure many quality metrics of output, such as factual correctness, naturalness, fluency, coherence, etc.
%Therefore, an automatic semantic similarity metric that can be learnt to capture syntactic, and semantic similarity, as well as other properties of generated text like quality, naturalness, fluency and coherence is necessary for evaluation of generated text.
% \jianfeng{note that this does not solve the problem of the dialog example, where the generated-output is semantically different from the reference.}
% RESOLVED
In this section we survey the NLG evaluation metrics that are computed using machine-learned models, with a focus on recent neural models.

\begin{table}[h]
\small
    \centering
    \begin{tabular}{|r|p{5cm}|p{5cm}|}
    \hline
         & Dialog Response Generation & Image Captioning \\
         \hline
    \multirow{5}{*}{Context}     & \textbf{Speaker A}: Hey John, what do you want to do tonight? & \multirow{5}{*}{\centering \includegraphics[height=2cm]{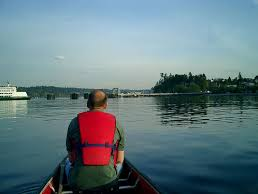}} \\
    & & \\
         & \textbf{Speaker B}: Why don’t we go see a movie? &  \\
          & & \\
         \hline
Ground-Truth & \textbf{Response:} Nah, I hate that stuff, let’s do something active. &  \textbf{Caption:} a man wearing a red life jacket is sitting in a canoe
on a lake\\
\hline
Model/Distorted Output & \textbf{Response:} Oh sure! Heard the film about Turing is out!  & \textbf{Caption:} a guy wearing a life vest is in a small boat on a
lake\\
\hline
\textsc{BLEU} & 0.0 & 0.20 \\
\hline
\textsc{ROUGE} & 0.0 & 0.57 \\
\hline
\textsc{WMD} & 0.0 & 0.10 \\
\hline
    \end{tabular}
    \caption{Demonstration of issues with using automatic evaluation metrics that rely on \textit{n}-gram overlap using two short-text generation tasks: dialog response generation and image captioning. The examples are adapted from \citet{liu-etal-2016-evaluate} and \citet{kilickaya_2016}.}
    \label{table:evalerrors}
\end{table}

%##################################
\subsection{Sentence Semantic Similarity Based Evaluation} 
\label{sssbe}
%##################################
Neural approaches to sentence representation learning seek to capture semantic meaning and syntactic structure of sentences from different perspectives and topics and to map a sentence onto an embedding vector using neural network models. 
As with word embeddings, NLG models can be evaluated by embedding each sentence in the generated and reference texts.
\begin{figure}[h!]
    \centering
    \includegraphics[width=1.0\textwidth]{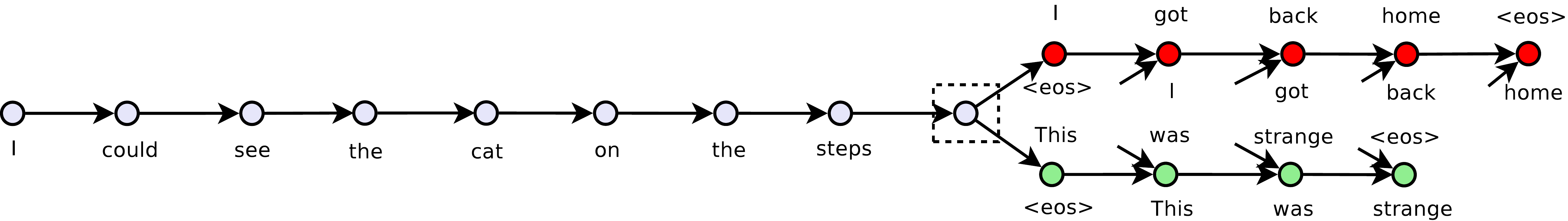}
    \caption{Illustration of Skip-Thoughts Vectors Model for sentence representation learning (Image Source: \citep{skipthought}).}
    \label{fig:skipthoughts}
\end{figure}

Extending word2vec \citep{word2vec} to produce word or phrase embeddings, one of the earliest sentence embeddings models, Deep Semantic Similarity Model (\textbf{\textsc{dssm}}) \citep{huang2013dssm} introduced a series of latent semantic models with a deep structure that projects two or more text streams (such as a query and multiple documents) into a common low-dimensional space where the relevance of one text towards the other text can be computed via vector distance.
The \textbf{\textsc{skip-thought}} vectors model \citep{skipthought} exploits the encoder-decoder architecture to
predict context sentences in an unsupervised manner (see Figure~\ref{fig:skipthoughts}). 
Skip-thought vectors allow us to encode rich contextual information by taking into account the surrounding context, but are slow to train. 
\textbf{\textsc{fastsent}} \citep{fastsent} makes training efficient by representing a sentence as the sum of its word embeddings, but also dropping any knowledge of word order. 
A simpler \textbf{\textsc{weighted sum}} of word vectors \citep{arora2017asimple} weighs each word vector by a factor similar to the tf-idf score, where more frequent terms are weighted less. 
Similar to \textsc{fastsent}, it ignores word order and surrounding sentences. % limiting the information that is encoded. 
Extending \textsc{dssm} models, \textbf{\textsc{infersent}} \citep{infersent} is an effective model, which uses \textsc{lstm}-based Siamese networks, with two additional advantages over the \textsc{fastsent}. It encodes word order and is trained on a high-quality sentence inference dataset.  
On the other hand, \textbf{\textsc{quick-thought}} \citep{logeswaran2018an} is based on an unsupervised model of universal sentence embeddings trained on consecutive sentences. Given an input sentence and its context, a classifier is trained to distinguish a context sentence from other contrastive sentences based on their embeddings. 

The recent large-scale pre-trained language models (PLMs) such as \textbf{\textsc{elmo}} and \textbf{\textsc{bert}} use contextualized word embeddings to represent sentences.
%, while the more powerful \textbf{\textsc{bert}} averages the last layer word representation of the model.
Even though these PLMs % recent neural contextual representation learning models 
outperform the earlier models such as \textsc{dssm}s, they are more computationally expensive to use for evaluating NLG systems.
% the generated text performance against human-generated reference text. 
For example, the sentence similarity metrics that use Transformer-based encoders, such as \textsc{bert} model \citep{bert} and its extension \textbf{\textsc{roberta}} \citep{roberta}, to obtain sentence representations
% (which is trained on a larger corpus than \textsc{bert}) 
are designed to learn textual similarities in sentence-pairs using distance-based similarity measures at the top layer as learning signal, such as cosine similarity similar to \textsc{dssm}. 
But both are much more computationally expensive than \textsc{dssm} due to the fact that they use a much deeper NN architecture, and need to be fine-tuned for different tasks.
% require both sentences to be fed into very deep NNs. 
To remedy this, \citet{reimers2019sentencebert} proposed
% a new \textsc{bert} model named 
\textbf{\textsc{sentbert}}, a fine-tuned \textsc{bert} on a ``general'' task to optimize the \textsc{BERT} parameters, so that a cosine similarity between two generated sentence embeddings is strongly related to the semantic similarity of the two sentences. 
Then the fine-tuned model can be used to evaluate various NLG tasks. 
Focusing on machine translation task, \textbf{\textsc{esim}} also computes sentence representations from \textsc{bert} embeddings (with no fine-tuning), and later computes the similarity between the translated text and its reference
using metrics such as the average recall of its reference \citep{chen-etal-2017-enhanced,mathur-etal-2019-putting}.

% #################################
\subsection{Regression-Based Evaluation}
\label{rbe}
\begin{figure}[h!]
    \centering
    \includegraphics[width=0.3\textwidth]{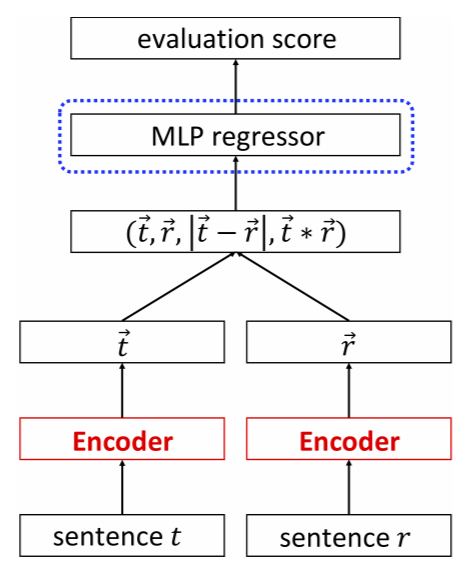}
    \caption{\small The sketch of the \textsc{ruse} metric. Image source \citep{logeswaran2018an}.}
    \label{fig:ruse}
\end{figure}

% #################################
\citet{shimanaka-etal-2018-ruse} proposed a segment-level machine translation evaluation metric named \textbf{\textsc{ruse}}. 
They treat the evaluation task as a regression problem to predict a scalar value to indicate the quality of translating a machine-translated hypothesis $t$ to a reference translation $r$. They first do a forward pass on the \textsc{GRU} (gated-recurrent unit) based on an encoder to generate $t$ and represent $r$ as a $d$-dimensional vector. Then, they apply different matching methods to extract relations between $t$ and $r$ by (1) concatenating $(\vec{t},\vec{r})$; (2) getting the element-wise product ($\vec{t}*\vec{r}$); (3) computing the absolute element-wise distance $|\vec{t}-\vec{r}|$ (see Figure ~\ref{fig:ruse}). 
\textsc{ruse} is demonstrated to be an efficient metric in machine translation shared tasks in both segment-level (how well the metric correlates with human
judgments of segment quality) and system-level (how well a given
metric correlates with the machine translation workshop official manual ranking) metrics.
% #################################
% \subsection{Distributional Divergence Estimation with Human Judgments}
\subsection{Evaluation Models with Human Judgments}
% #################################
For more creative and open-ended text generation tasks, such as chit-chat dialog, story generation, or online review generation, current evaluation methods are only useful to some degree. As we mentioned in the beginning of this section, word-overlap metrics are ineffective as there are often many plausible references in these scenarios and collecting them all is impossible. Even though human evaluation methods are useful in these scenarios for evaluating aspects like coherency, naturalness, or fluency, aspects like diversity or creativity may be difficult for human judges to assess as they have no knowledge about the dataset that the model is trained on \citep{hashimoto-etal-2019-unifying}. 
Language models can learn to copy from the training dataset and generate samples that a human judge will rate as high in quality, but may fail in generating diverse samples (i.e., samples that are very different from training samples), as has been observed in social chatbots \citep{li-etal-2016-diversity,zhou2020design}.
%diversity metric when more
%as all the generated samples can be found in the training data. 
%As we discussed in the previous sections,
A language model optimized only for perplexity may generate coherent but bland responses. 
%more diverse sentences, but will not necessarily be coherent or of good quality.\footnote{Training a language model on a large corpus for a long period of time would improve the model's skills in generating coherent and grammatical sentences indistinguishable from human \citep{radford2019language}.}. 
Such behaviours are observed when generic pre-trained language models are used for downstream tasks `as-is' without fine-tuning on in-domain datasets of related downstream tasks. A commonly overlooked issue is that conducting human evaluation for every new generation task can be expensive and not easily generalizable.  

To calibrate human judgments and automatic evaluation metrics, model-based approaches that use human judgments as attributes or labels have been proposed. \cite{adem2017} introduced a model-based evaluation metric, \textsc{adem}, which is learned from human judgments for dialog system evaluation, specifically response generation in a chatbot setting.
Using Twitter data (each tweet response is a reference, and its previous dialog turns are its context), they have different models (such as RNNs, retrieval-based methods, or other human responses) generate responses and ask humans to judge the appropriateness of the generated response given the context. 
For evaluation they use a higher quality labeled Twitter dataset \citep{ritter2011data-driven}, which contains dialogs on a variety of topics.  
\begin{figure}[h!]
    \centering
    \includegraphics[width=0.95\textwidth]{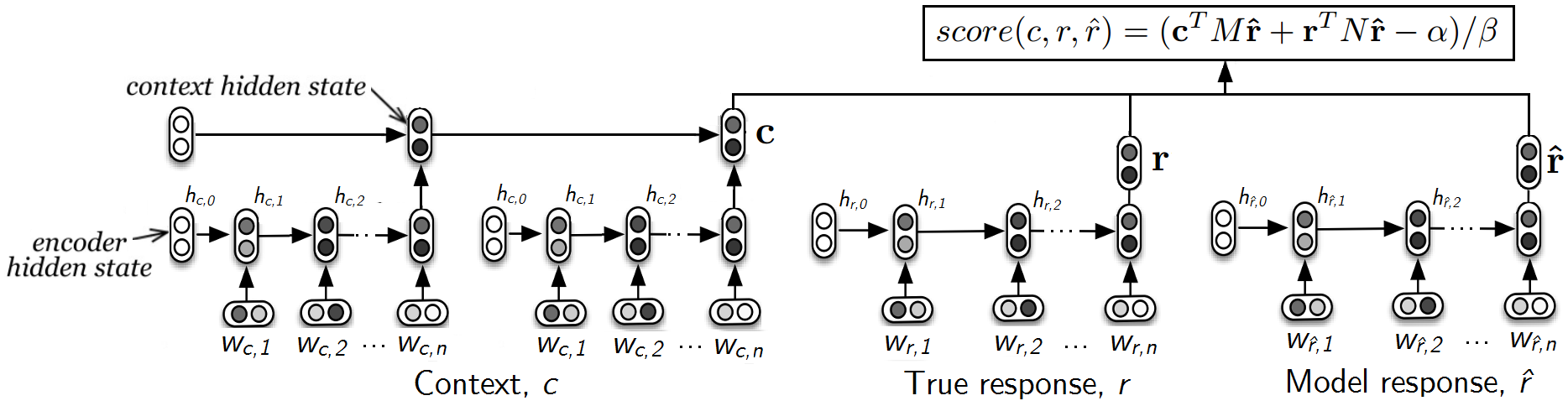}
    \caption{\small The ADEM evaluation model. Image source \citep{adem2017}.}
    \label{fig:adem}
\end{figure}

Using this score-labeled dataset, the \textsc{adem} evaluation model is trained as follows: 
First, a latent variational recurrent encoder-decoder model (\textsc{vhred}) \citep{vhred} is pre-trained on a dialog dataset to learn to represent the context of a dialog. \textsc{vhred} encodes the dialog context into a vector representation, from which the model generates samples of initial vectors to condition the decoder model to generate the next response. Using the pre-trained \textsc{vhred} model as the encoder, they train \textsc{adem} as follows (see Figure~\ref{fig:adem}). First, the dialog context, $c$, the model generated response $\hat{r}$, and the reference response $r$ are fed to \textsc{vhred} to get their embedding vectors, $\mathbf{c}$, $\mathbf{\hat{r}}$ and $\mathbf{r}$. Then, each embedding is linearly projected so that the model response $\mathbf{\hat{r}}$ can be mapped onto the spaces of the dialog context and the reference response to calculate a similarity score. The similarity score measures how close the model responses are to the context and the reference response after the projection, as follows:

\begin{equation}
score(c, \hat{r}, r)= 
(\textbf{c}^TM\mathbf{\hat{r}} + \mathbf{r}^TN\mathbf{\hat{r}}-\alpha)/\beta
\end{equation}

\textsc{adem} is optimized for squared error loss between the predicted score and the human judgment score with L-2 regularization in an end-to-end fashion. 
The trained evaluation model is shown to correlate well with human judgments.
\textsc{adem} is also found to be conservative and give lower scores to plausible responses.

\begin{figure}[h!]
    \centering
    \includegraphics[width=0.5\textwidth]{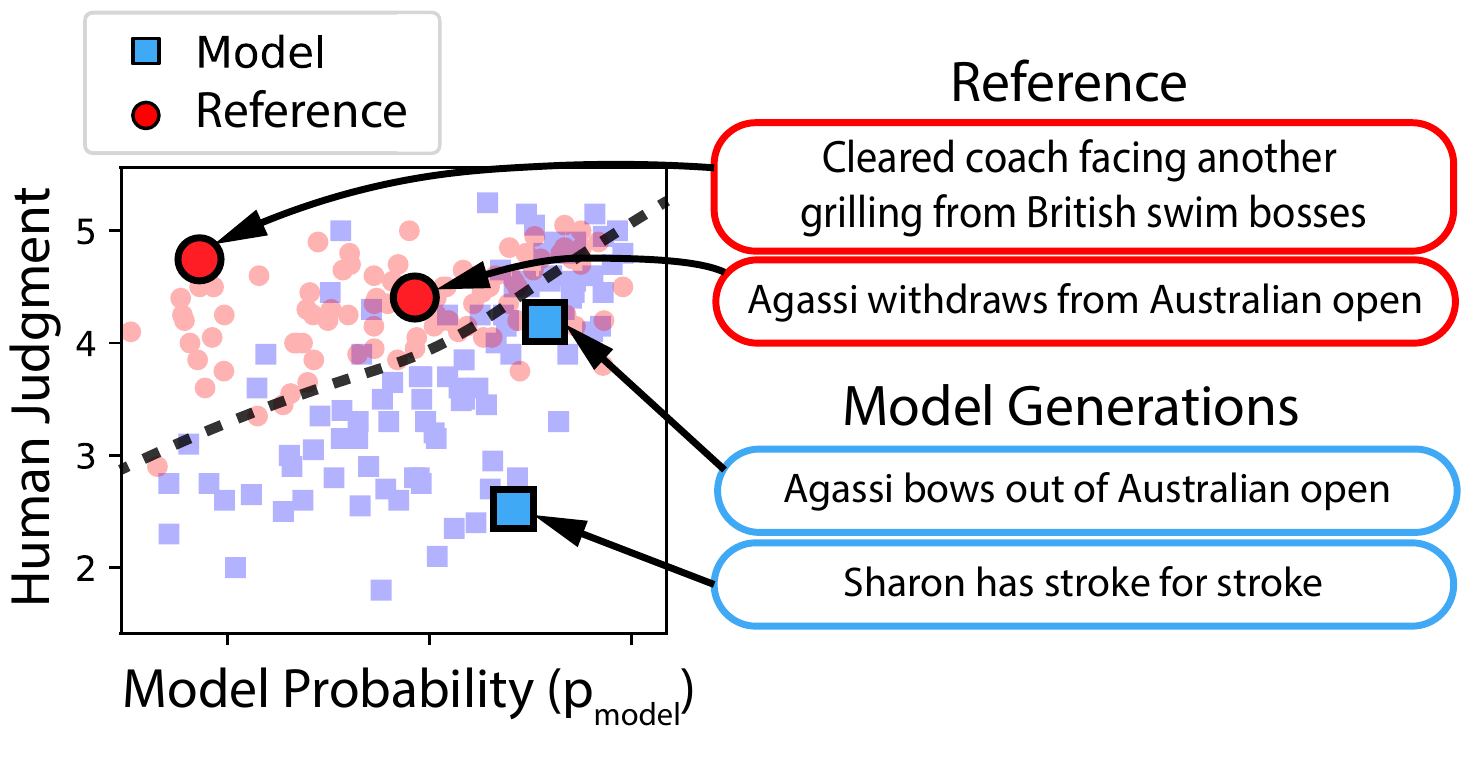}
    \caption{\small \textsc{huse} can
identify samples with defects in quality (\textit{Sharon has
stroke for stroke}) and diversity (\textit{Cleared coach facing}). Image Source: \citep{hashimoto-etal-2019-unifying}.}
    \label{fig:huse}
\end{figure}

With the motivation that a good evaluation metric should capture both the quality and the diversity of the generated text, \citet{hashimoto-etal-2019-unifying} proposed a new evaluation metric named Human Unified with Statistical
Evaluation (\textbf{\textsc{huse}}), which focuses on more creative and open-ended text generation tasks, such as dialog and story generation. 
Unlike the \textsc{adem} metric, which relies on human judgments for training the model, \textsc{huse} combines statistical evaluation and human evaluation metrics in one model, as shown in Figure~\ref{fig:huse}.

\textsc{huse} considers the conditional generation task that, given a context $x$ sampled from a prior distribution $p(x)$, outputs a distribution over possible sentences $p_{model}(y|x)$. 
The evaluation metric is designed to determine the similarity of the output distribution $p_{model}$ and a human generation reference distribution $p_{ref}$. 
This similarity is scored using an \textit{optimal discriminator} that determines whether a sample comes from the reference or hypothesis (model) distribution (Figure~\ref{fig:huse}). 
For instance, a low-quality text is likely to be sampled from the model distribution. 
The discriminator is implemented approximately using two probability measures: (i) the probability of a sentence under the model, which can be estimated using the text generation model, and (ii) the probability under the reference distribution, which can be estimated based on human judgment scores. %Leveraging both the model probabilities and human judgments ensures to evaluate the quality and diversity of the generated text against reference. 
On summarization and chitchat dialog tasks, \textsc{huse} has been shown to be effective to detect low-diverse generations that humans fail to detect.

% #################################
\subsection{BERT-Based Evaluation} 
% #################################
Given the strong performance of \textsc{bert} \citep{bert} across many tasks, 
%over many transformer-based encoders in the last year, 
there has been work that uses \textsc{bert} or similar pre-trained language models for evaluating NLG tasks, such as summarization and dialog response generation. 
Here, we summarize some of the recent work that fine-tunes \textsc{bert} to use as evaluation metrics for downstream text generation tasks. 
% \jianfeng{Note that BERT-based metrics are embedding-based metrics, described in Sec. 3.2.2.}

\begin{figure}[h!]
    \centering
    \includegraphics[width=1.0\textwidth]{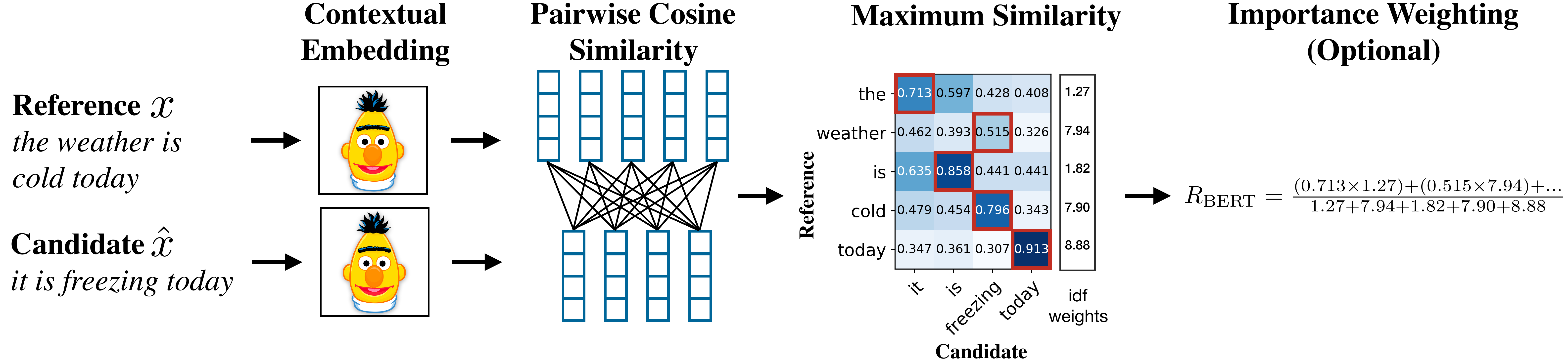}
    \caption{\small Illustration of \textsc{bertscore} metric. Image Source: \citet{bert-score}.}
    \label{fig:bertscore}
\end{figure}

One of the \textsc{bert}-based models for semantic evaluation is \textbf{\textsc{bertscore}} \citep{bert-score}. As illustrated in Figure~\ref{fig:bertscore}, it leverages the pre-trained contextual embeddings from \textsc{bert} and matches words in candidate and reference sentences by cosine similarity. 
\textsc{bertscore} has been shown to correlate well with human judgments on sentence-level and system-level evaluations. Moreover, \textsc{bertscore} computes precision, recall, and F1 measures, which are useful for evaluating a range of NLG tasks.

\citet{kan2019neural} presented a \textsc{bert}-based evaluation method called \textbf{\textsc{roberta-sts}} 
to detect sentences that are logically contradictory or unrelated, regardless whether they are grammatically plausible. 
Using \textsc{roberta} \citep{roberta} as a pre-trained language model, \textsc{roberta-sts} is fine-tuned on the STS-B dataset \citep{STBDataset} to learn the similarity of sentence pairs on a Likert scale. 
Another evaluation model is fine-tuned on the Multi-Genre Natural Language Inference Corpus \citep{N18-1101} in a similar way to learn to predict logical inference of one sentence given the other. 
Both model-based evaluators, \textsc{roberta-sts} and its extension, have been shown to be more robust and correlate better with human evaluation than automatic evaluation metrics such as \textsc{bleu} and \textsc{rouge}. 
% even when the inputs are corrupted. 

\begin{figure}[h]
    \centering
    \includegraphics[width=0.5\textwidth]{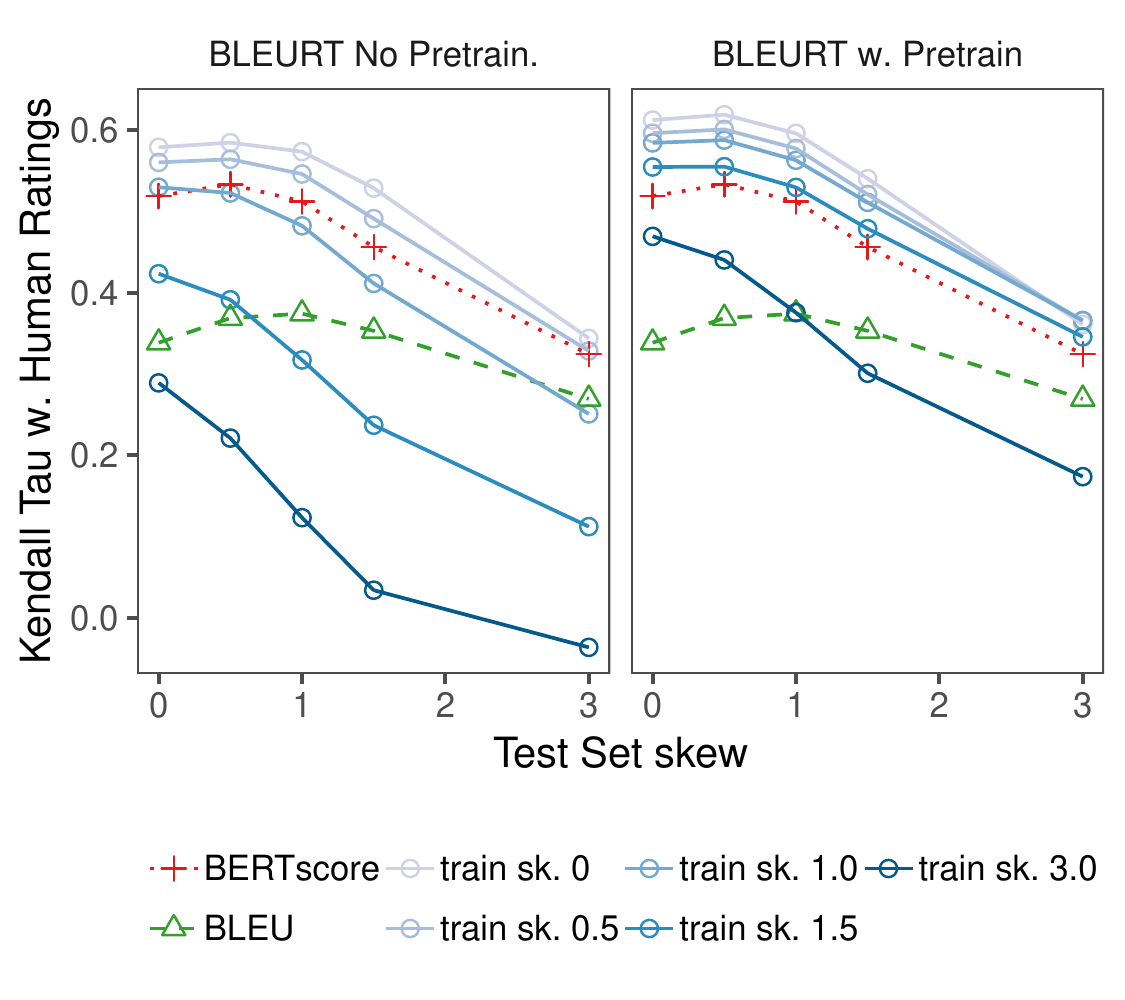}
    \caption{\small Agreement between \textsc{bleurt} and human
ratings for different skew factors in train and test. Image Source: \citet{bleurt}}
    \label{fig:bleurt}
\end{figure}

Another recent \textsc{bert}-based machine-learned evaluation metric is \textbf{\textsc{bleurt}} \citep{bleurt}, which was proposed to evaluate various NLG systems. 
%This metric is based on \textsc{bert}, and the recipe to calculate 
The evaluation model is trained as follows: A checkpoint from \textsc{bert} is taken and fine-tuned on synthetically generated sentence pairs using automatic evaluation scores such as \textsc{bleu} or \textsc{rouge}, and then further fine-tuned on system-generated outputs and human-written references using human ratings and automatic metrics as labels. The fine-tuning of \textsc{bleurt} on synthetic pairs is an important step because it improves the robustness to quality drifts of generation systems.  
As shown in the plots in Figure~\ref{fig:bleurt}%\liz{broken ref}
, as the NLG task gets more difficult, the ratings get closer as it is easier to discriminate between “good” and “bad” systems than to rank “good” systems.
To ensure the robustness of their metric, they investigate with training datasets with different characteristics, such as when the training data is highly skewed or out-of-domain. 
They report that the training skew has a disastrous effect on
\textsc{bleurt} without pre-training; this pre-training makes \textsc{bleurt} significantly more robust to quality drifts.

As discussed in Section~\ref{human}, humans can efficiently evaluate the performance of two models side-by-side, and most embedding-based similarity metrics reviewed in the previous sections are based on this idea. 
Inspired by this, the \textbf{comparator evaluator} \citep{zhou2020learning} was proposed to evaluate NLG models by
learning to compare a pair of generated sentences by
fine-tuning \textsc{bert}. A text pair relation classifier is trained to compare the task-specific quality of a sample hypothesis and reference based on the win/loss rate. 
Using the trained model, a skill rating system is built. This system is similar to the player-vs-player games in which the players are evaluated by observing a record of wins and losses of multiple players. Then, for each player, the system infers the value of a latent, unobserved skill variable that indicates the records of wins and losses. 
On story generation and 
open domain dialogue response generation tasks, the comparator evaluator metric demonstrates high correlation with human evaluation.
% #################################
\begin{figure}[t]
    \centering
    \includegraphics[width=0.7\textwidth]{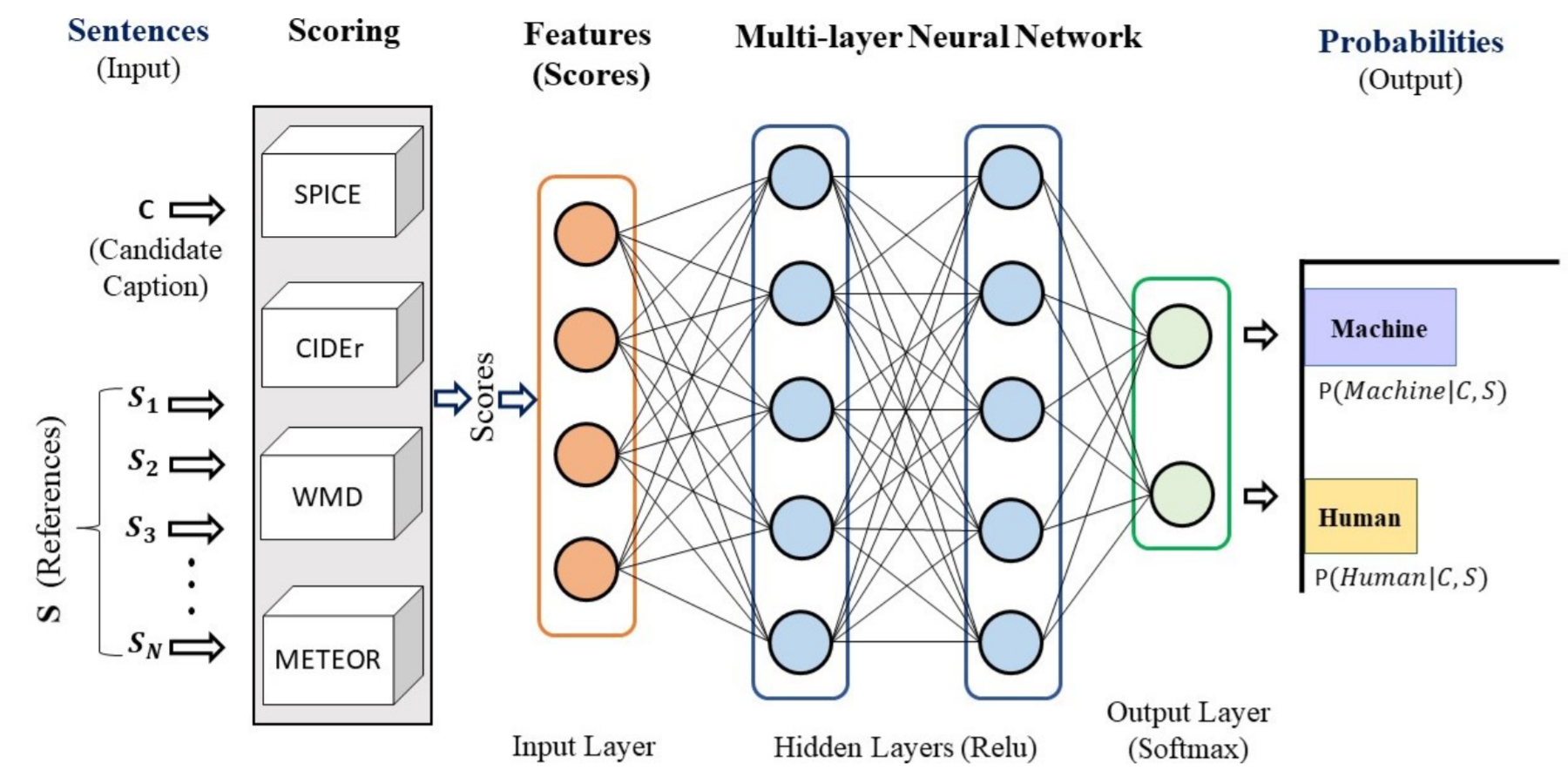}
    \caption{\small Composite Metrics model architecture. Image Source: \citep{sharif-etal-2018-learning}.}
    \label{fig:composite}
\end{figure}

% #################################
\subsection{Evaluating Factual Correctness}
\label{factevals}
% #################################
An important issue in text generation systems is that the model's generation could be factually inconsistent, caused by distorted or fabricated facts about the source text. Especially in document summarization tasks, the models that abstract away salient aspects, have been shown to generate text with up to 30\% factual inconsistencies \citep{kryciski2019evaluating,falke-etal-2019-ranking,zhu2020boosting}. There has been a lot of recent work that focuses on building models to verify the factual correctness of the generated text, focusing on semantically constrained tasks such as document summarization or image captioning, some of which we summarize here. 
\begin{figure}[h!]
    \centering
    \includegraphics[width=0.8\textwidth,trim={0.5cm 6cm 1cm 2cm},clip]{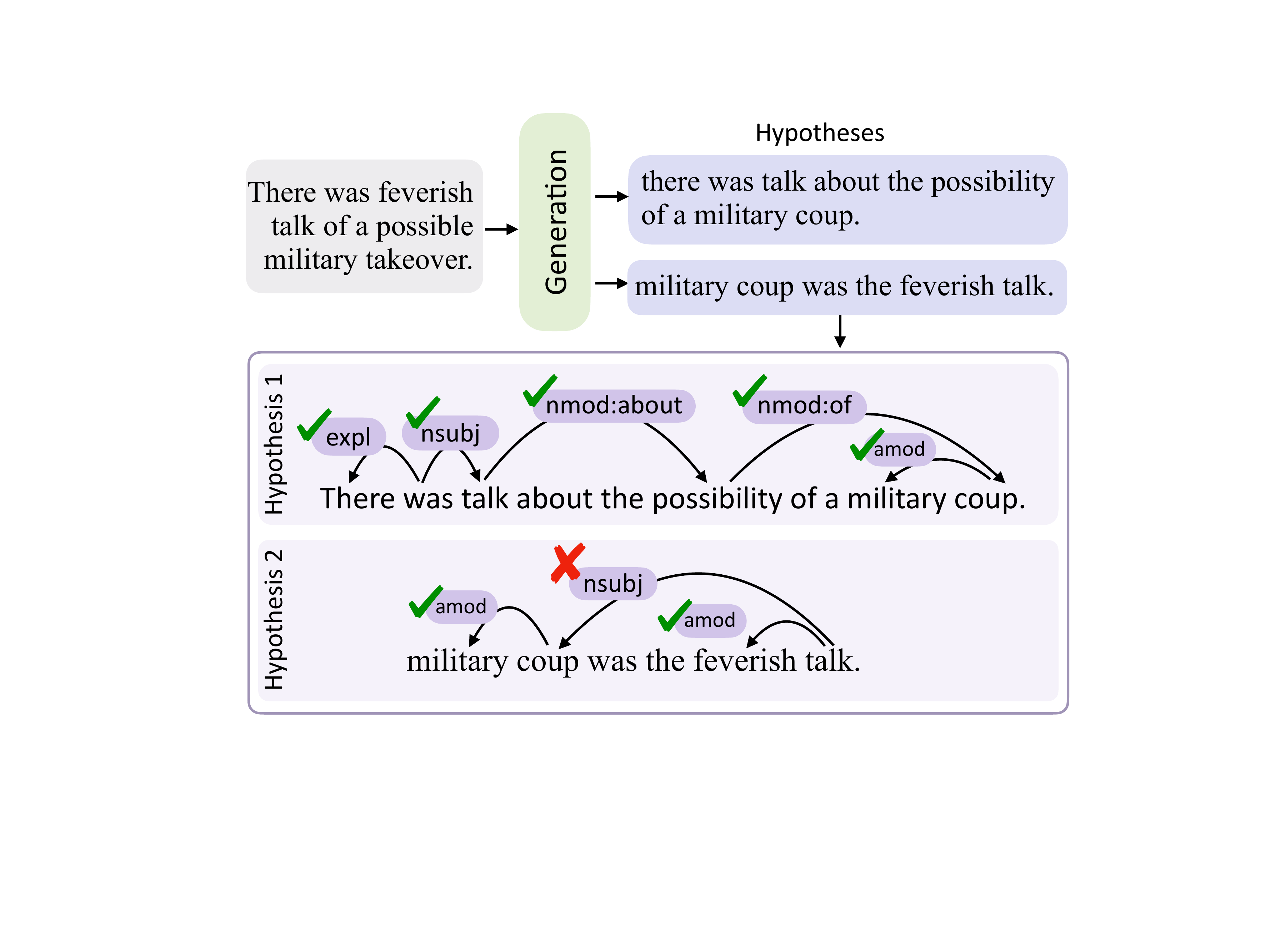}
    \caption{\small Illustration of dependency arc entailment formulation using a filtered set of Stanford Enhanced Dependencies. Image Source: \citep{goyal-durrett-2020-evaluating}.}
    \label{giyal_durett_fig}
\end{figure}

Some recent evaluation metrics have addressed factual correctness via entailment-based models \citep{falke-etal-2019-ranking,Maynez2020OnFA,dusek-kasner-2020-evaluating}. However, these sentence-level, entailment-based approaches do not capture which part of the generated text is non-factual. \citeauthor{goyal-durrett-2020-evaluating} presented a new localized entailment-based approach using dependency trees to reformulate the entailment problem at the dependency arc level. Specifically, they align the the semantic relations yielded by the dependency arcs (see Figure \ref{giyal_durett_fig}) in the generated output summary to the input sentences. Their dependency arc entailment model improves factual consistency and shows stronger correlations with human judgments in generation tasks such as summarization and paraphrasing. 

Models adhering to the facts in the source have started to gain more attention in ``conditional'' or ``grounded'' text generation tasks, such as document summarization \citep{kryciski2019evaluating} and data-to-text generation \citep{ehud_data_to_text07,Lebret2016,Sha2017,Puduppully18,wang-2019-revisiting,nan2021dart}. 
In one of the earlier works on structured data-to-text generation, \citet{wiseman2017challenges} dealt with the coherent generation of multi-sentence summaries of tables or database records. In this work, they first trained an auxiliary model as relation extraction classifier (entity-mention pairs) based on information extraction to evaluate how well the text generated by the model can capture the information in a discrete set of records.
Then the factual evaluation is based on the alignment between the entity-mention predictions of this classifier against the source database records. Their work was limited to a single domain (basketball game tables and summaries) and assumed that the tables has similar attributes, which can be limiting for open-domain data-to-text generation systems. 

\cite{dhingra2019handling} extended this approach and introduced the \textsc{parent} measure. Their evaluation approach first aligns the entities in the table and the reference and generated text with a neural attention-based model and later measures similarities on word overlap, entailment and other metrics over the alignment. They conduct a large scale human evaluation study which yielded that \textsc{parent} correlates with human judgments better than several n-gram match and information extraction based metrics they used for evaluation. 
\citeauthor{parikh2020totto} proposed a new controllable text generation task, \textsc{totto}, which generates a sentence to describe a highlighted cell in a given table and extended the \textsc{parent} to adapt to their tasks so the metric takes into account the highlighted cell in the table.

Factual consistency evaluations have also appeared in multi-modal generation tasks, such as image captioning. In one such work \citep{chen2018factual}, a new style-focused factual \textsc{rnn}-type decoder is constructed to allow the model to preserve factual information in longer sequences without requiring additional labels.
In this model, they query a reference model to adaptively learn to add factual information into the model.

% \label{factevals}
\begin{figure}[h!]
    \centering
    \includegraphics[width=1.0\textwidth]{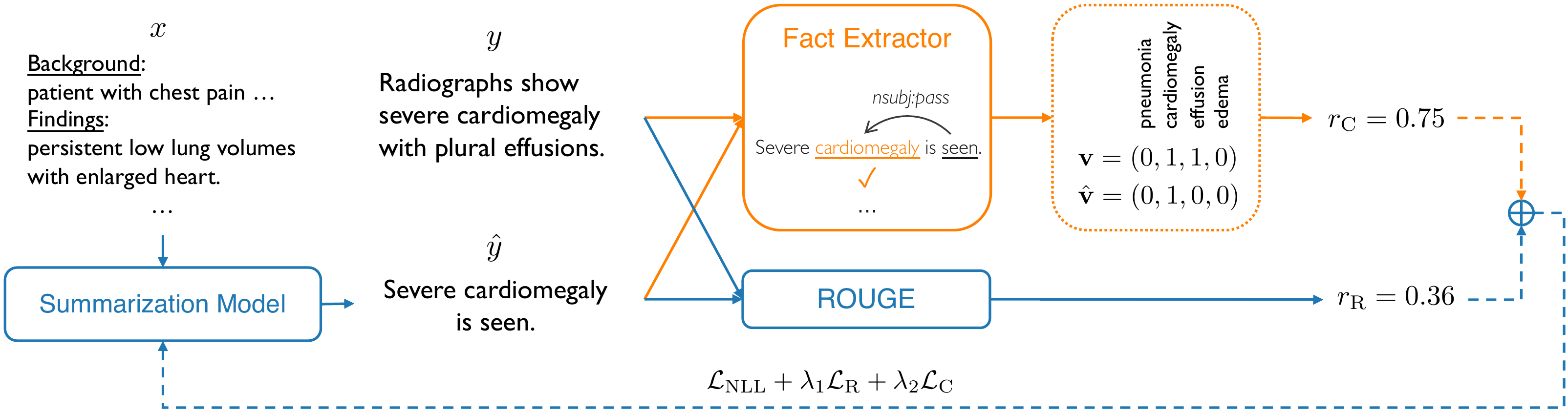}
    \caption{\small Illustration of the training strategy of the factually correct summarization model. Image Source: \citep{factsum}.}
    \label{fig:factsum}
\end{figure}
\citet{factsum} proposed a way to tackle the problem of factual correctness in summarization models. Focusing on summarizing radiology reports, they extend pointer networks for abstractive summarization by introducing a reward-based optimization that trains the generators to obtain more rewards when they generate summaries that are factually aligned with the original document. Specifically, they design a fact extractor module so that the factual accuracy of a generated summary can be measured and directly optimized as a reward using policy gradient, as shown in Figure~\ref{fig:factsum}. This fact extractor is based on an information extraction module and extracts and represents the facts from generated and reference summaries in a structured format. 
The summarization model is updated
via reinforcement learning using a combination of the \textsc{NLL} (negative log likelihood) loss, a \textsc{rouge}-based loss, and a factual correctness-based loss (\textit{Loss}=$\mathcal{L}_{NLL}$+$\lambda_1\mathcal{L}_{rouge}$+$\lambda_2\mathcal{L}_{fact}$). Their work suggests that for domains in which generating factually correct text is crucial, a carefully implemented information extraction system
can be used to improve the factual correctness of
neural summarization models via reinforcement learning.

\begin{figure}[h!]
    \centering
    \includegraphics[width=0.3\textwidth]{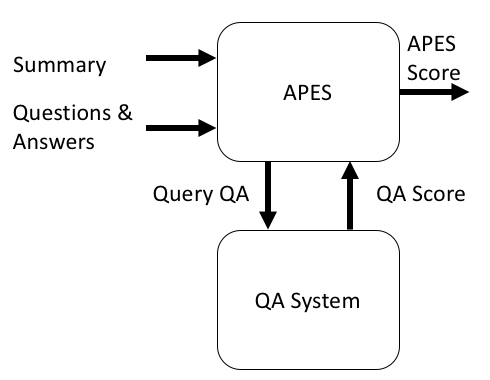}
    \caption{APES evaluation flow. Image Source: \citep{hashimoto-etal-2019-unifying}.}
    \label{fig:apes}
\end{figure}

To evaluate the factual consistency of the text generation models, \citet{eyaletal2019question} presented a question-answering-based parametric evaluation model named Answering Performance for
Evaluation of Summaries (\textsc{apes}) (see Figure~\ref{fig:apes}). Their evaluation model is designed to evaluate document summarization and is based on the hypothesis that
the quality of a generated summary is associated with the 
number of questions (from a set of relevant ones)
that can be answered by reading the summary. 

To build such an evaluator to assess the quality of generated summaries, they introduce two components: 
(a) a set of relevant questions for each source document and (b) a question-answering system.
They first generate questions from each reference summary by masking each of the 
named entities present in the reference based on the method described in \citet{teachingmachines}.
For each reference summary, this results in several  
triplets in the form \textit{(generated summary, question, answer)}, where \textit{question} refers to 
%as named entities present in the reference summary
the sentence containing the masked entity, 
\textit{answer} refers to the masked entity, and 
the \textit{generated summary} is generated by their summarization model. Thus,
for each generated summary, metrics can be derived based on the accuracy of the question answering system in retrieving
the correct answers from each of the associated
triplets. This metric is useful for summarizing documents for domains that contain lots of named entities, such as biomedical or news article summarization. 

% #################################
\subsection{Composite Metric Scores} 
% #################################
The quality of many NLG models like machine translation and image captioning can be evaluated for multiple aspects, such as adequacy, fluency, and diversity.
Many composite metrics have been proposed to capture a multi-dimensional sense of quality.
\citet{sharif-etal-2018-learning} presented a machine-learned composite metric for evaluating image captions. 
The metric incorporates a set of existing metrics such as \textsc{meteor}, \textsc{wmd}, and \textsc{spice} to
measure both adequacy and fluency. 
They evaluate various combinations of the metrics they chose to compose and and show that their composite metrics correlate well with human judgments.

\begin{figure}[t]
    \centering
    \includegraphics[width=0.7\textwidth]{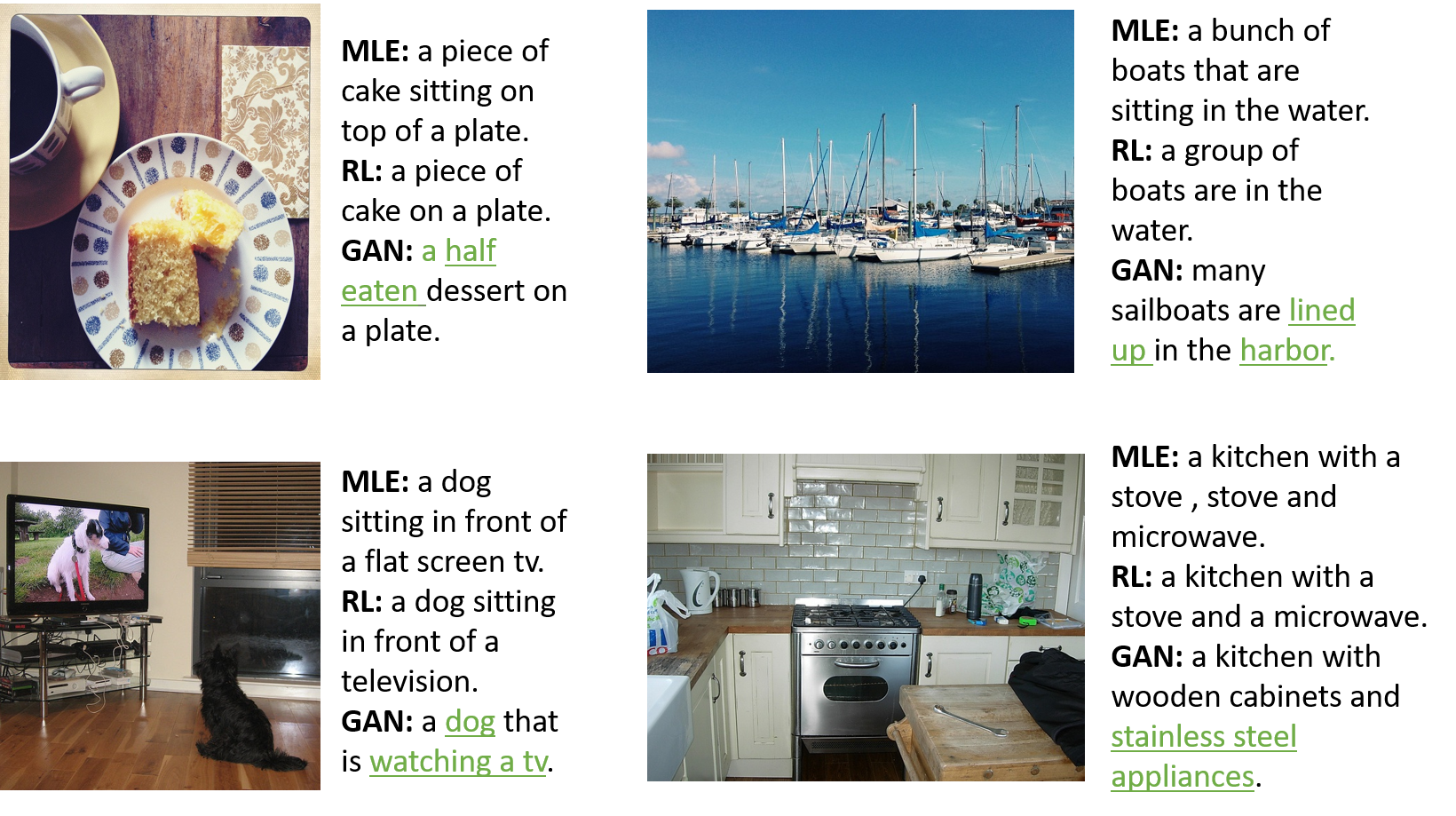}\\
    \includegraphics[width=0.7\textwidth]{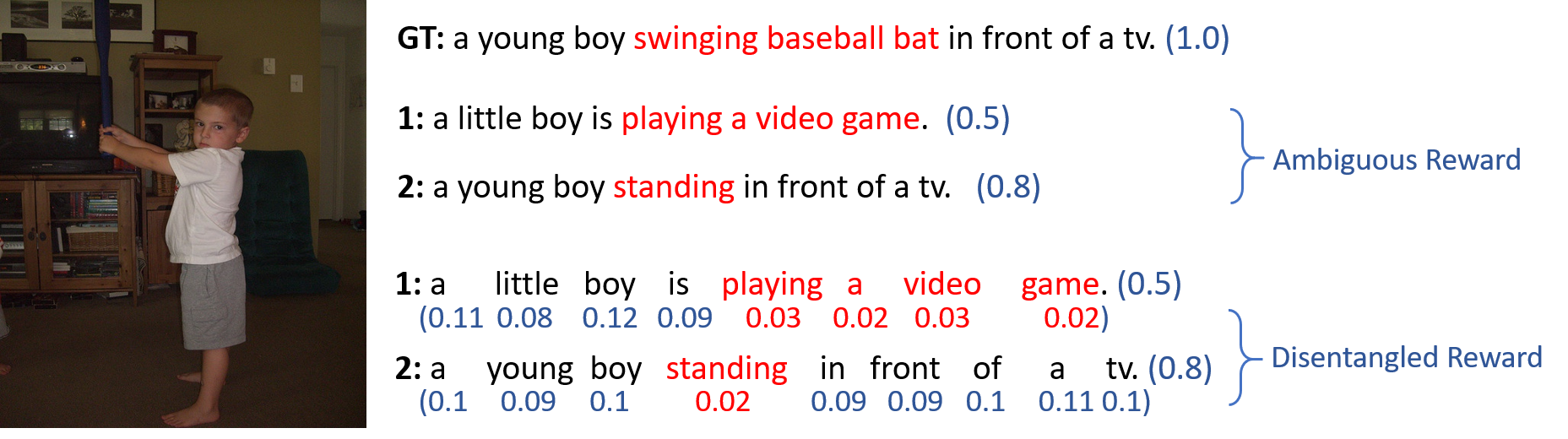}
    \caption{\small (Top four images) Example image captions using different learning objectives: MLE: maximum likelihood learning, GAN: Generative Adversarial Networks, RL: Reward-based reinforcement learning. (Bottom image) Example generations from Adversarial Inverse Reinforcement Learning (rAIRL). Image Source: \citep{li2020learning}.}
    \label{fig:composite2}
\end{figure}

%aspects of assessment and improves upon the baseline metrics that focus on one or the other. 
\citet{li2020learning} propose a composite reward function to evaluate the performance of image captions. 
The approach is based on refined Adversarial Inverse Reinforcement Learning (rAIRL), which eases the reward ambiguity (common in reward-based generation models) by decoupling the reward for each word in a sentence. 
The proposed composite reward is shown on MS COCO data to achieve state-of-the-art performance on image captioning. 
Some examples generated from this model that uses the composite reward function are shown in Figure~\ref{fig:composite2}. They have shown that their metric not only generates grammatical captions but also correlates well with human judgments.

\section{Shared Tasks for NLG Evaluation}
\label{sharedtasks}
Shared tasks in NLG are designed to boost the development of sub-fields and continuously encourage researchers to improve upon the state-of-the-art. With shared tasks, the same data and evaluation metrics are used to efficiently benchmark models. 
NLG shared tasks are common not only because language generation is a growing research field with numerous unsolved research challenges, but also because many NLG generation tasks do not have an established evaluation pipeline.
NLG researchers are constantly proposing new shared tasks as new datasets and tasks are introduced to support efficient evaluation of novel approaches in language generation.
Even though shared tasks are important for NLG research and evaluation, there are potential issues that originate from the large variability and a lack of standardisation in the organisation of shared tasks, not just for language generation but for language processing in general. In \citep{parra-escartin-etal-2017-ethical}, some of these ethical concerns are discussed. 

In this section we survey some of the shared tasks that focus on the evaluation of text generation systems that are aimed at comparing and validating different evaluation measures.

% ########################
\subsection{Generating Referring Expressions}
% ########################
The Attribute Selection for Generating Referring Expressions (\textbf{GRE}) (\textsc{asgre}) Challenge \citep{ASREG} was one of the first shared-task evaluation challenges in NLG. It was designed for the content determination of the GRE task, selecting the properties to describe an intended referent. The goal of this shared task was to evaluate the submitted systems on \textit{minimality} (the proportion of descriptions in the system-generated output that are maximally brief compared to the original definition),\textit{uniqueness} and \textit{humanlikeness}. 

% ########################
\subsection{Embedded Text Generation}
% ########################
To spur research towards human-machine communication in situated settings, Generating Instructions in Virtual Environments (\textbf{GIVE}) has been introduced as a challenge and an evaluation testbed for NLG \citep{koller-etal-2009-software}. 
In this challenge a human player is given a task to solve in a simulated 3D space.
A generation module’s task is to guide the human player, using natural language instructions. 
Only the human user can effect any changes in the world, by moving around, manipulating objects, etc.
This challenge evaluates NLG models on referring expression generation, aggregation, grounding, realization, and user modeling.
This challenge has been organized in four consecutive years \citep{koller-etal-2011-software}.

% ########################
\subsection{Regular Expression Generation (REG) in Context}
% ########################
The goal in this task is to map a representation of an intended referent in a given textual context to a full surface form. The representation of the intended referring expression maybe one from possible list of referring expressions for that referent and/or a set of semantic and syntactic properties. This challenge has been organized under different sub-challenges: \textbf{GREC-Full} has focused on improving the referential clarity and fluency of the text in which systems were expected to replace regular expressions and where necessary to produce as clear, fluent and coherent a text as possible \citep{belzkow2010grec}. The \textbf{GREC-NEG} Task at Generation Challenges 2009 \citep{belz-etal-2009-grec-named} evaluated models in select correct coreference chains for all people entities mentioned in short encyclopaedic texts about people collected from Wikipedia.

% ########################
\subsection{Regular Expression Generation from Attribute Sets}
% ########################
This task tries to answer the following question: Given a symbol corresponding to an intended referent, how do we work out the semantic content of a referring expression that uniquely identifies the entity in question? \citep{regas}. 
The input to these models consists of sets of attributes (e.g., \{type=lamp, colour=blue, size=small\}), where at least one attribute set is labelled the intended referent, and the remainder are the distractors. Then the task is to build a model that can output a set of attributes for the intended referent that uniquely distinguishes it from the distractors. \citet{gatt-etal-2008-tuna} have introduced the \textbf{\textsc{tune} Corpus} and the \textbf{\textsc{tuna} Challenge} based on this corpus that covered a variety of tasks, including attribute selection for referring expressions, realization and end-to-end referring expression generation. 

% ########################
\subsection{Deep Meaning Representation to Text (SemEval)}
% ########################
\textbf{SemEval} is a series of NLP workshops organized around the goal of advancing the current state of the art in semantic analysis and to help create high-quality annotated datasets to approach challenging problems in natural language semantics. Each year a different shared task is introduced for the teams to evaluate and benchmark models. For instance, Task 9 of the SemEval 2017 challenge was \citep{semeval-2017-international} on text generation from AMR (Abstract Meaning Representation), which has focused on generating valid English sentences given AMR  \citep{banarescu-etal-2013-abstract} annotation structure.

% ########################
\subsection{WebNLG}
% ########################
The \textbf{WebNLG} challenge introduced a text generation task from RDF triples to natural language text, providing a corpus and common benchmark for comparing the microplanning capacity of the generation systems that deal with resolving and using referring expressions, aggregations, lexicalizations, surface realizations and sentence segmentations \citep{gardent-etal-2017-webnlg}. A second challenge has taken place in 2020 \citep{zhou-lampouras-2020-webnlg}, three years after the first one, in which the dataset size increased (as did the coverage of the verbalisers) and more categories and an additional language were included to promote the development of knowledge extraction tools, with a task that mirrors the verbalisation task.

\subsection{E2E NLG Challenge} 
Introduced in 2018, E2E NLG Challenge \citep{dusek-etal-2018-findings} provided a high quality and large quantity training dataset for evaluating response generation models in spoken dialog systems. It introduced new challenges such that models should jointly learn sentence planning and surface realisation, while not requiring costly alignment between meaning representations and corresponding natural language reference texts. 

\subsection{Data-to-Text Generation Challenge}
Most existing work in data-to-text (or table-to-text) generation
focused on introducing datasets and benchmarks rather than organizing challanges. Some of these earlier works include: \textsc{eathergov} \citep{liang-etal-2009-learning}, \textsc{robocup} \citep{robo},
\textsc{rotowire} \citep{wiseman2017challenges}, \textsc{e2e} \citep{novikova-etal-2016-crowd}, \textsc{wikibio} \citep{lebret-etal-2016-neural} and recently \textsc{totto} \citep{parikh2020totto}.
\citeauthor{banik-etal-2013-kbgen} introduced a text generation from knowledge base\footnote{http://www.kbgen.org} challenge in 2013 to benchmark
various systems on the content realization stage of generation. Given a set of relations which
form a coherent unit, the task is to generate complex sentences that are grammatical and fluent in English.

\subsection{GEM Benchmark}
Introduced in ACL 2021, the \textsc{gem} benchmark\footnote{https://gem-benchmark.com} \citep{Gehrmann2021TheGB} aims to measure the progress in NLG, while continuously adding new datasets, evaluation metrics and human evaluation standards. \textsc{gem} provides an environment by providing easy testing of different NLG tasks and evaluation strategies.

\section{Examples of Task-Specific NLG Evaluation}
\label{applications}

In the previous sections, we reviewed a wide range of NLG evaluation metrics individually. 
However, these metrics are constantly evolving due to rapid progress in more efficient, reliable, scalable and sustainable neural network architectures for training neural text generation models, as well as ever growing compute resources. Nevertheless, it is not easy to define what really is an ``accurate,'' ``trustworthy'' or even ``efficient'' metric for evaluating an NLG model or task. 
Thus, in this section we present how these metrics can be jointly used in research projects to more effectively evaluate NLG systems for real-world applications. 
We discuss two NLG tasks, automatic document summarization and long-text generation, that are sophisticated enough that multiple metrics are required to gauge different aspects of the generated text's quality.

\begin{table*}[th]
    \centering
    \small
    \begin{tabular}{p{0.46\linewidth} | p{0.46\linewidth}}
        Summarization Evaluation Metrics & Long-Text Generation Evaluation Metrics \\
        \hline
        \textsc{rouge}, \textsc{bleu}, \textsc{f-score}, \textsc{sera}, ... & \textsc{rouge}, \textsc{bleu}, \textsc{f-score}, ... \\
        model-based factual correctness metrics & entity based evaluation \\
        Q/A based factuality metrics & syntactic measures for writing style \\
        human-based evaluations & human-based evaluations \\
        \hline
\end{tabular}
    \caption{Metrics mentioned in each example text generation project.}
    \label{tab:case}
\end{table*}

% #####################################
\subsection{Automatic Document Summarization Evaluation}
% #####################################
A text summarization system aims to extract useful content from a reference document and generate a short summary that is coherent, fluent, readable, concise, and consistent with the reference document. 
There are different types of summarization approaches, which can be grouped by their tasks into (i) \textbf{generic} text summarization for broad topics; (ii) \textbf{topic-focused} summarization, e.g., a scientific article, conversation, or meeting summarization; and (iii) \textbf{query-focused} summarization, such that the summary answers a posed query.
These approaches can also be grouped by their method: (i) \textbf{extractive}, where a summary is composed of a subset of sentences or words in the input document; and (ii) \textbf{abstractive}, where a summary is generated on-the-fly and often contains text units that do not occur in the input document. 
Depending on the number of documents to be summarized, these approaches can also be grouped into single-document or multi-document summarization.

\begin{figure}[h]
    \centering
    \includegraphics[width=0.7\textwidth]{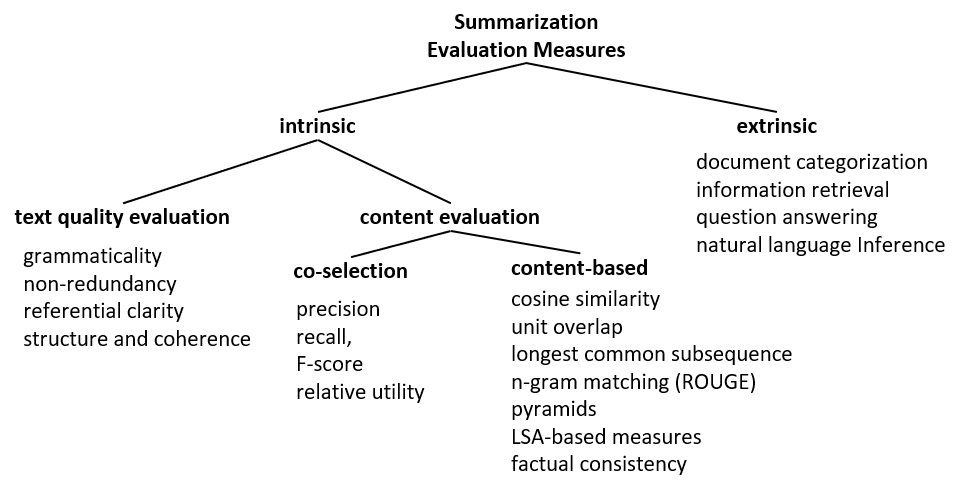}
    \caption{Taxonomy of summarization evaluation methods. Extended from  \citet{sumeval09}.}
    \label{fig:sumeval}
\end{figure}

Evaluation of text summarization, regardless of its type, measures the system's ability to generate a summary based on: (i) a set of criteria that are not related to references \citep{Referenceless2017}, (ii) a set of criteria that measure its closeness to the reference document, or (iii) a set of criteria that measure its closeness to the reference summary. 
Figure~\ref{fig:sumeval} shows the taxonomy of evaluation metrics \citep{sumeval09} in two categories: intrinsic and extrinsic, which will be explained below.

\subsubsection{Intrinsic Methods}
Intrinsic evaluation of generated summaries can focus on the generated text's content, text quality, and factual consistency, each discussed below.

\paragraph{Content.} 
%The most commonly used criteria, 
Content evaluation compares a generated summary to a reference summary using automatic metrics.
% such as \textit{n}-gram overlap.  
The most widely used metric for summarization is \textsc{rouge}, though
%is widely used for summarization evaluation task, 
other metrics, such as \textsc{bleu} and \textsc{f-score}, are also used. 
%It has been shown that 
Although \textsc{rouge} has been shown to correlate well with human judgments for generic text summarization, the correlation is lower for topic-focused summarization like extractive meeting summarization \citep{meetingsumrouge}. 
Meetings are transcripts of spontaneous speech, and thus usually contain disfluencies, such as pauses (e.g., `um,' `uh,' etc.), discourse markers (e.g., `you know,' `i mean,' etc.), repetitions, etc. 
%and the data comes with utterances which are labeled with speaker information. 
\cite{meetingsumrouge} find that after such disfluencies are cleaned, the \textsc{rouge} score is improved. They even observed fair amounts of improvement in the correlation between the \textsc{rouge} score and human judgments when they include the speaker information of the extracted sentences from the source meeting to form the summary.

\paragraph{Quality.} Evaluating generated summaries based on quality has been one of the challenging tasks for summarization researchers. As basic as it sounds, since the definition of a ``good quality summary'' has not been established and finding the most suitable metrics to evaluate quality remains an open research area. 
Below are some criteria of text, which are used in recent papers as human evaluation metrics to evaluate the quality of generated text in comparison to the reference text.
%Nevertheless, among various criteria to compare the summarization system performance proposed by recent work are
\begin{itemize}
    \item \textbf{Coherence} and \textbf{Cohesion} measure how clearly the ideas are expressed in the summary \citep{lapatadiscourse2}. In particular, the idea that, in conjunction with cohesion, which is to hold the context as a whole, coherence should measure how well the text is organised and ``hangs together.'' Consider the examples in Table~\ref{table:big_wall_of_examples},
    %we include two example text outputs
    from the scientific article abstract generation task. The models must include factual information, but it must also be presented in the right order to be coherent.
   
    \item \textbf{Readability and Fluency}, associated with non-redundancy, are linguistic quality metrics used to measure how repetitive %much of the same ideas are repeated in 
    the generated summary is and how many spelling and grammar errors there are in the generated summary %include good spelling and grammar 
    \citep{lapatadiscourse1}. 
    \item \textbf{Focus} indicates how many of the main ideas of the document are captured, while avoiding superfluous details.
    \item \textbf{Informativeness}, which is mostly used to evaluate question-focused summarization, measures how well the summary answers a question. Auto-regressive generation models trained to generate a short summary text given a longer document(s) may yield shorter summaries due to reasons relating to bias in the training data or type of the decoding method (e.g., beam search can yield more coherent text compared to top-k decoding but can yield shorter text if a large beam size is used) \citep{Huang_2017}. Thus, in comparing different model generations, the summary text length has also been used as an informativeness measure since a shorter text typically preserves less information \citep{sumeval16}.

\end{itemize}

\begin{table*}[t] %496/500
\centering
    \small 
  \begin{tabular}{p{2.15cm}|p{12.55cm}}
  \hline 
    \multirow{4}{*}{\textbf{Gold}} & We investigate mutual benefits between syntax and semantic roles using neural network models, by studying a parsing->SRL pipeline, a SRL->parsing pipeline, and a simple joint model by embedding sharing. The integration of syntactic and semantic features gives promising results in a Chinese Semantic Treebank... \\ \hline
    \multirow{3}{*}{\textbf{PGen}} & In this paper, we propose a novel approach to learn \textit{\textbf{syntactic and semantic role labeling models}} \textit{to semantic role labeling (wsd)}. \textit{In the first} neural network models induce non-linear \textit{feature features} from word and \textit{\textbf{part-of-speech (pos) parsing}}. We show that semantic features can be used to learn...  \\ \hline
    \multirow{3}{*}{\textbf{\textbf{Co-OpNet}}}& In this paper, we explore the use of neural network models to jointly train \textit{\textbf{semantic role labelers}} and \textit{\textbf{parsers}} for \textbf{\textit{semantic role labeling (srl)}}. We first propose a simple neural \textit{\textbf{srl}} model that uses a neural \textbf{\textit{long shortterm memory (lstm)-based parser}} to represent the output of an \textit{\textbf{srl}} system...\\
    \hline
  \end{tabular}
  \caption{Example of gold and generated abstracts from baseline Pointer Networks + Coverage~\cite{see-etal-2017-get} (\textbf{PGen}) and a recently proposed model, \textbf{Co-OpNet}~\citep{saadia_each2021}, in the NLP scientific domain. Coherence issues and factual errors in generated abstracts are highlighted in \textit{italics}. Correct terminology and transitional phrases that contribute to coherent flow are also highlighted by properly delineating sections of abstracts in \textbf{\textit{bold}} and \textbf{\textit{italics}}. (Table Source: \citep{saadia_each2021}.}
  \label{table:big_wall_of_examples}
\end{table*}

These quality criterion are widely used as evaluation metrics for human evaluation in document summarization. They can be used to compare a system-generated summary to a source text, a human-generated summary, or to another system-generated summary.

% and Theautomatic evaluation.

% ###################################
\paragraph{Factual Consistency.} 
% ###################################

One thing that is usually overlooked in document summarization tasks is evaluating the generated summaries' factual correctness. 
It has been shown in many recent work on summarization that models frequently generate factually incorrect text. This is partially because the models are not trained to be factually consistent and can generate about anything related to the prompt,  Table~\ref{tab:summarization-errors} shows a sample summarization model output, in which the claims made are not consistent with the source document \citep{kryciski2019evaluating}. \citeauthor{zhang-etal-2020-optimizing}  

It is imperative that the summarization models are factually consistent and that any conflicts between a source document and its generated summary (commonly referred to as \textit{faithfulness} \citep{durmus-etal-2020-feqa,Wang_2020}) can be easily measured, especially for domain-specific summarization tasks like patient-doctor conversation summarization or business meeting summarization. 
As a result, 
%the summarization research community has been seeing a new spike in 
factual-consistency-aware and faithful text generation research has drawn a lot of attention in the community in recent years \citep{kryscinski-etal-2019-neural,kryciski2019evaluating,factsum,wang2020asking,durmus-etal-2020-feqa,Wang_2020}. 
A common approach is to use a model-based approach, in which a separate component is built on top of a summarization engine that can evaluate the generated summary based on factual consistency, as discussed in Section~\ref{factevals}.

    \begin{table*}[t]
        \begin{center}
        \resizebox{\linewidth}{!}{%
        \small
        \begin{tabular}{p{0.48\linewidth}|p{0.48\linewidth}} 
        \hline
        \multicolumn{2}{l}{\textbf{Source article fragments}} \\ 
        \hline
        (CNN) \textcolor{teal}{The mother of a quadriplegic man who police say was left in the woods for days cannot be extradited} to face charges in Philadelphia until she completes an unspecified ``treatment," Maryland police said Monday. The Montgomery County (Maryland) Department of Police took \textcolor{teal}{Nyia Parler, 41}, into custody Sunday (...) &
        (CNN) The classic video game ``Space Invaders" was developed in Japan \textcolor{teal}{back in the late 1970's} -- and now their real-life counterparts are the topic of an earnest political discussion in Japan's corridors of power. Luckily, Japanese can sleep soundly in their beds tonight as the government's top military official earnestly revealed that (...) \\
        \hline
        \multicolumn{2}{l}{\textbf{Model generated claims}} \\
        \hline
        \textcolor{red}{Quadriplegic man} Nyia Parler, 41, \textcolor{red}{left in woods for days} can not be extradited. &
        Video game ``Space Invaders" was developed in Japan back \textcolor{red}{in 1970}. \\
        \hline
        \end{tabular}
        }
        \caption{
        Examples of factually incorrect claims output by summarization models.
        Green text highlights the support in the source documents for the generated claims; red text highlights the errors made by summarization models. Table Source~\citep{kryciski2019evaluating}.
        }
        \label{tab:summarization-errors}
        \end{center}
    \end{table*}

\subsubsection{Extrinsic Summarization Evaluation Methods}
Extrinsic evaluation metrics test
the generated summary text by how it impacts the performance of  downstream tasks, 
%These metrics are mostly used in downstream tasks 
such as relevance assessment, reading comprehension, and question answering. 
\citet{assesrelevance} propose a new metric, \textsc{sera} (Summarization Evaluation by Relevance Analysis), for summarization evaluation based on the content relevance of the generated summary and the human-written summary. 
They find that this metric yields higher correlation with human judgments compared to \textsc{rouge}, especially on the task of scientific article summarization. 
\citet{Eyal_2019} and \citet{wang2020asking} 
% use reading-comprehension techniques to 
measure the performance of a summary by using it to answer a set of questions regarding the salient entities in the source document. 

% #####################################
\subsection{Long Text Generation Evaluation}
\begin{figure*}
\centering
\includegraphics[page=1,trim={0 14in 0 2.in}, clip,width=\textwidth]{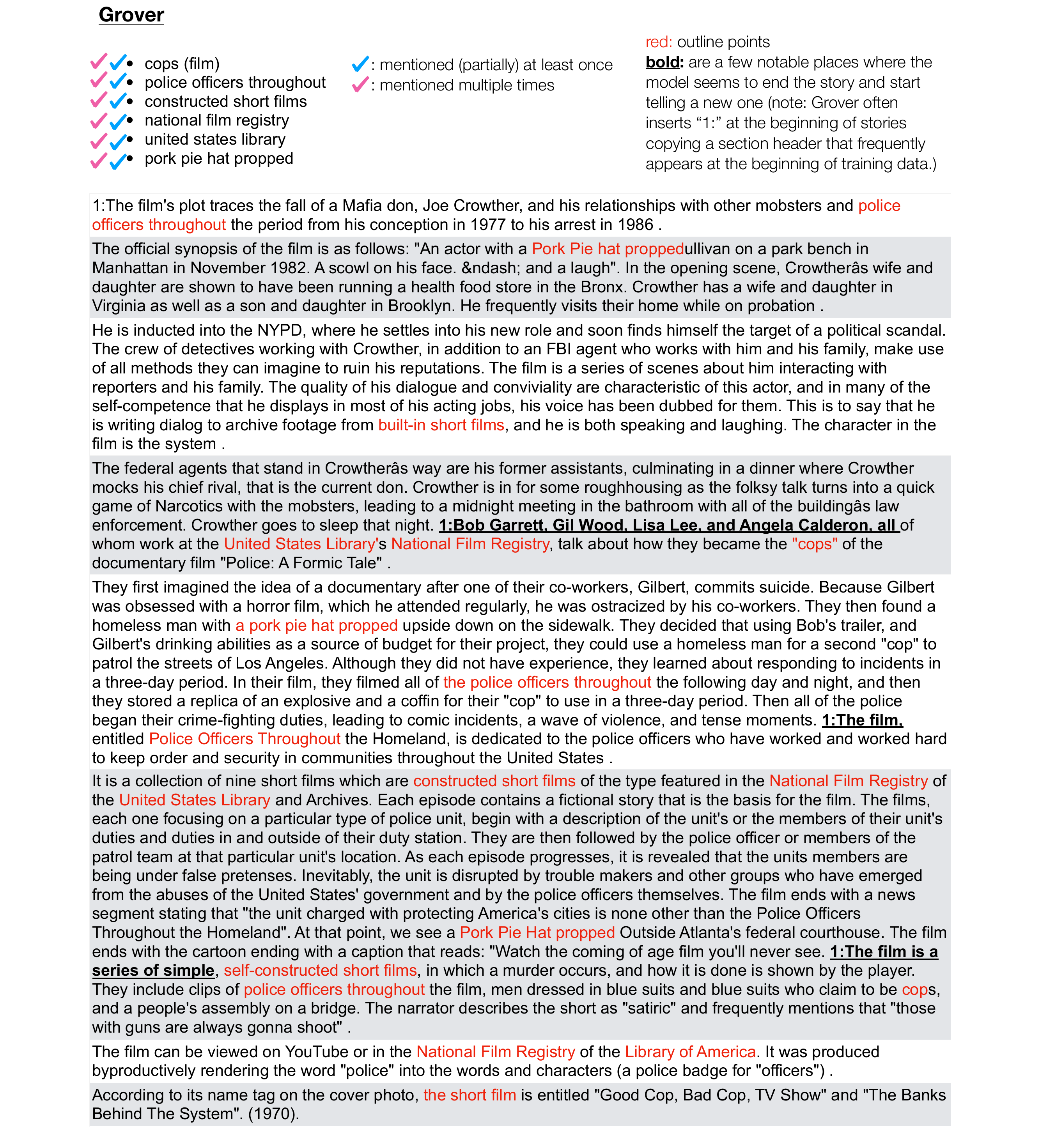}
    \caption{Example document generated using \grover{}. Red text indicates where plot points are mentioned, while bold text marks the beginning of a new story. (Table Source \citep{plotmachines}.)}
    \label{fig:gen:grover2}
\end{figure*}

\begin{figure*}
\centering
\includegraphics[page=2,trim={0 18in 0 2.in}, clip,width=\textwidth]{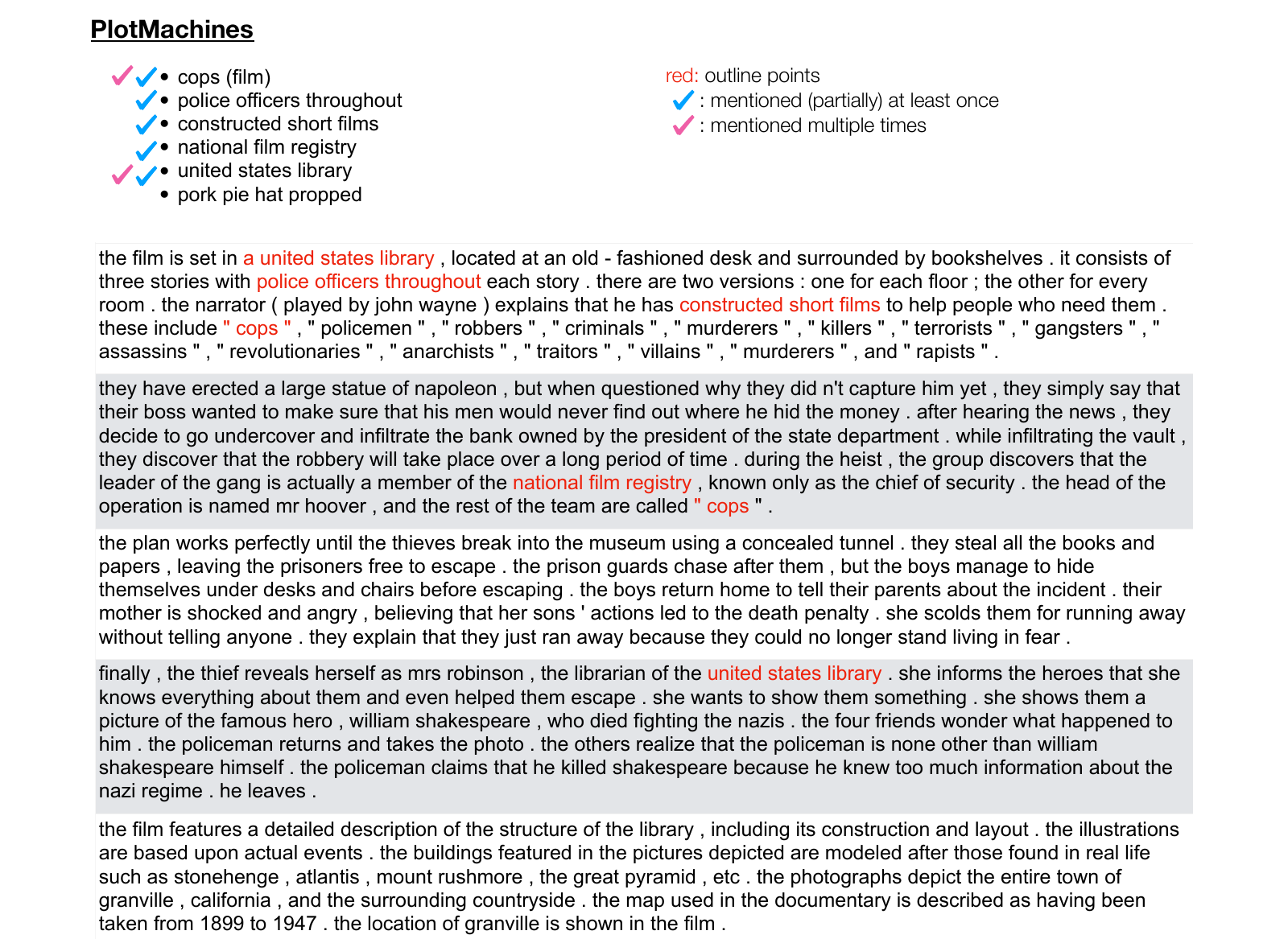}
    \caption{Example document generated using \modelname{}. Red text indicates where plot points are mentioned. (Table Source \citep{plotmachines}.)}
    \label{fig:gen:pm2}
\end{figure*}

% #####################################
A long text generation system aims to generate multi-sentence text, such as a single paragraph or a multi-paragraph document. Common applications of long-form text generation are document-level machine translation, story generation, news article generation, poem generation, summarization, and image description generation, to name a few.  
This research area presents a particular challenge to state-of-the-art approaches that are based on statistical neural models, which are proven to be insufficient to generate coherent long text. 
As an example, in Figure~\ref{fig:gen:grover2} and \ref{fig:gen:pm2} we show two  generated text from two long-text generation models, \grover{} \citep{grover} and \modelname{} \citep{plotmachines}. Both of these controlled text models are designed to generate a multi-paragraph story given a list of attributes (in these examples a list of outline points are provided), and the models should generate a coherent long story related to the outline points. These examples demonstrate some of the cohesion issues with these statistical models. For instance, in the \grover{} output, the model often finishes the story and then starts a new story partway through the document.  In contrast, \modelname{} adheres more to a beginning-middle-ending structure.
For example, GPT-2 \citep{gpt} 
can generate remarkably fluent sentences, and even paragraphs, for a given topic or a prompt. 
However, as more sentences are generated and the text gets longer, it starts to wander, switching to unrelated topics and becoming incoherent \citep{plotmachines}. 

Evaluating long-text generation is a challenging task. New criteria need to be implemented to measure the quality of long generated text, such as inter-sentence or inter-paragraph coherence in language style and semantics.   
Although human evaluation methods are commonly used, we focus our discussion on automatic evaluation methods in this section.

%################################### 
\subsubsection{Evaluation via Discourse Structure} 
%################################### 
 Text with longer context (e.g., documents, longer conversations, debates, movies scripts, etc.) usually consist of sections (e.g., paragraphs, sets, topics, etc.) that constitute some structure, and in natural language generation such structures are referred to as \textit{discourse} \citep{jurafskydiscourse}. 
 Considering the discourse structure of the generated text is crucial in evaluating the system. Especially in open-ended text generation, % in which a language model is asked to generate text given a prompt, 
 the model needs to determine the topical flow, structure of entities and events, and their relations in a narrative flow that is coherent and fluent. One of the major tasks in which discourse plays an important role is document-level machine translation \citep{Gong2015}.  \cite{Hajlaoui2013} present a new metric called Accuracy of Connective Translation (ACT) \citep{meyer2012} that uses a combination of rules and automatic metrics to compare the discourse connection between the source and target documents. \cite{Joty2017}, on the other hand, compare the source and target documents based on the similarity of their discourse trees. 

%################################### 
\subsubsection{Evaluation via Lexical Cohesion} 
%################################### 
Lexical cohesion is a surface property of text and refers to the way textual units are linked together
grammatically or lexically. 
Lexical similarity \citep{lapatadiscourse2} is one of the most commonly used metrics in story generation. 
\cite{Roemmele2017} filter the $n$-grams based on lexical semantics and only use adjectives, adverbs, interjections, nouns, pronouns, proper nouns, and verbs for lexical similarity measure. 
Other commonly used metrics compare reference and source text on word- \citep{word2vec} or sentence-level \citep{skipthought} embedding similarity averaged over the entire document. Entity co-reference is another metric that has been used to measure coherence \citep{Elsner2008}. An entity should be referred to properly in the text and should not be used before introduced. \citet{Roemmele2017} capture the proportion of the entities in the generated sentence that are co-referred to an entity in the corresponding context as a metric of entity co-reference, in which a higher co-reference score indicates higher coherence.

In machine translation, \cite{maruf2019survey} introduce a feature that can identify lexical cohesion at the sentence level via word-level clustering using WordNet \citep{wordnet} and stemming to obtain a score for each word token, which is averaged over the sentence. 
They find that this new score improves correlation of \textsc{bleu} and \textsc{ter} with human judgments. 
Other work, such as \cite{Gong2015}, uses topic modeling together with automatic metrics like \textsc{bleu} and \textsc{meteor} to evaluate lexical cohesion in machine translation of long text. 
\cite{chow-etal-2019-wmdo} investigate the position of the word tokens in evaluating the \textit{fluency} of the generated text. They modify \textsc{wmd} by adding a fragmentation penalty to measure the fluency of a translation for evaluating machine translation systems.

%################################### 
\subsubsection{Evaluation via Writing Style} 
%################################### 
%Writing style has been mainly used in literature to distinguish successful writers from unsuccessful ones. 
\cite{lingstyle} show that an author's writing is consistent in style across a particular work. 
Based on this finding, \cite{Roemmele2017} propose to measure the quality of generated text based on whether it presents a consistent writing style. %follows the same style. 
They capture the category distribution of individual words between the story context and the generated following sentence using their part-of-speech tags of words (e.g., adverbs, adjectives, conjunctions, determiners, nouns, etc.). 

Text style transfer reflects the creativity of the generation model in generating new content. 
Style transfer can help rewrite a text in a different style, which is useful in creative writing such as poetry generation \citep{poetrygenration}. One metric that is commonly used in style transfer is the classification score obtained from a pre-trained style transfer model \citep{styletransfer}. 
This metric measures whether a generated sentence has the same style as its context. %in several long text generation tasks. 

%################################### 
\subsubsection{Evaluation with Multiple References} 
%################################### 
One issue of evaluating text generation systems is the diversity of generation, especially when the text to evaluate is long. The generated text can be fluent, valid given the input, and informative for the user, but it still may not have lexical overlap with the reference text or the prompt that was used to constrain the generation. 
This issue has been investigated extensively \citep{li-etal-2016-diversity,diversityandquality,nucleussampling,welleck2019neural,Gao2018}. 
%A solution to improve the robustness of evaluation metrics is by 
Using multiple references that cover as many plausible outputs as possible is an effective solution to improving the correlation of automatic evaluation metrics (such as adequacy and fluency) with human judgments, as demonstrated in machine translation \citep{mtsurveyhan2018,mtevalroadmapsurvey} and other NLG tasks.

\section{Conclusions and Future Directions}
\label{conclusion}

Text generation is central to many NLP tasks, including machine translation, dialog response generation, document summarization, etc.
%and vision-language tasks such as image captioning and visual story generation. 
With the recent advances in neural language models, the research community has made significant progress in developing new NLG models and systems for challenging tasks like multi-paragraph document generation or visual story generation. %translating a document to another language.
%have become more efficient and of a quality that is useful to people. 
With every new system or model comes a new challenge of evaluation. 
This paper surveys the NLG evaluation methods in three categories:
%that have been recently developed. 

\begin{itemize}
    \item \textbf{Human-Centric Evaluation.} Human evaluation is the most important for developing NLG systems and is considered the gold standard when developing automatic metrics. But it is expensive to execute, and the evaluation results are difficult to reproduce.

    \item \textbf{Untrained Automatic Metrics.} %As discussed in Chapter~\ref{metric}, 
    Untrained automatic evaluation metrics are widely used to monitor the progress of system development.

    A good automatic metric needs to correlate well with human judgments. For many NLG tasks, it is desirable to use multiple metrics to gauge different aspects of the system's quality.

    \item \textbf{Machine-Learned Evaluation Metrics.} In the cases where the reference outputs are not complete, we can train an evaluation model to mimic human judges. However, as pointed out in \citet{Gao2018}, any machine-learned metrics might lead to potential problems such as overfitting and `gaming of the metric.'
 
    \end{itemize}

We conclude this paper by summarizing some of the challenges of evaluating NLG systems:

\begin{itemize}
    
    \item \textbf{Detecting machine-generated text and fake news.}
    %Checking and Verifiability.}
    As language models get stronger 
    by learning from increasingly larger corpora of human-written text, they can generate text that is not easily distinguishable from human-authored text. Due to this, 
    %fake article detection system builders are developing 
    new systems and evaluation methods have been developed to detect if a piece of text is  machine- or human-generated. A recent study \citep{Schuster2019} reports the results of a fact verification system to identify inherent bias in training datasets that cause fact-checking issues. In an attempt to combat fake news, \cite{factcheckingtweets} present an extensive analysis of tweets and a new tweet generation method to identify fact-checking tweets (among many tweets), which were originally produced to persuade posters to stop tweeting fake news. 
    \citeauthor{gehrmann-etal-2019-gltr} introduced \textsc{GLTR}, which is a tool that helps humans to detect if a text is written by a human or generated by a model.  
    Other research focuses on factually correct text generation, with a goal of providing users with accurate information. \cite{massarelli2019decoding} introduce a new approach for generating text that is factually consistent with the knowledge source. \cite{kryciski2019evaluating} investigate methods of checking the consistency of a generated summary against the document from which the summary is generated. \citet{grover} present a new controllable language model that can generate an article with respect to a given headline, yielding more trustworthy text than human-written text of fake information. 
    Nevertheless, large-scale language models (even controllable ones), have a tendency to hallucinate and generate nonfactual information, which the model designers should measure and prevent. Future work should focus on the analysis of the text generated from large-scale language models, emphasize careful examination of such models in terms of how they learn and reproduce
    potential biases that in the training data \citep{sheng-etal-2020-towards, bender_2021}.

    \item \textbf{Making evaluation explainable.} 
    Explainable AI refers to AI and machine learning methods that can provide human-understandable justifications for their behaviour \citep{Ehsan2019}. Evaluation systems that can provide reasons for their decisions are beneficial in many ways. For instance, 
    %if the verification of the factual consistency of the text generator is explained, 
    the explanation could help system developers to identify the root causes of the system's quality problems such as unintentional bias, repetition, or factual inconsistency. % or why system malfunctions happen. 
    The field of explainable AI is growing, particularly in generating explanations of classifier predictions in NLP tasks \citep{Ribeiro2016,Ribeiro2018AnchorsHM,tokelevelexp}. Text generation systems that use evaluation methods that can provide justification or explanation for their decisions will be more trusted by their users. Future NLG evaluation research should focus on developing easy-to-use, robust, and explainable evaluation tools.

    \item \textbf{Improving corpus quality.}  Creating high-quality datasets with multiple reference texts is essential for not only improving the reliability of evaluation but also for allowing the development of new automatic metrics that correlate well with human judgments 
    \citep{belz-reiter-2006-comparing}. Among many important critical aspects of building corpora for natural language generation tasks, the \textit{accuracy}, \textit{timeliness}, \textit{completeness}, \textit{cleanness} and \textit{unbiasedness} of the data plays a very important role. The collected corpus (whether created manually or automatically through retrieval or generation) must be accurate so the generation models can serve for the downstream tasks more efficiently. The corpora used for language generation tasks should be relevant to the corresponding tasks so intended performance can be achieved. Missing information, information biased toward certain groups, ethnicities, religions, etc. could prevent the models from gathering accurate insights and could damage the efficiency of the task performance \citep{eckart-etal-2012-influence,mcguffie2020radicalization,barbaresi2015,bender_2021,Gehrmann2021TheGB}.

    \item \textbf{Standardizing evaluation methods.} Most untrained automatic evaluation metrics %that are used to evaluate NLG task performance today 
    are standardized using open source platforms like Natural Language Toolkit (NLTK)\footnote{nltk.org} or spaCy\footnote{spacy.io}. Such platforms can significantly simplify the process of benchmarking different models. However, there are still many NLG tasks that use task-specific evaluation metrics, such as metrics to evaluate the contextual quality or informativeness of generated text.
    There are also no standard criteria for human evaluation methods for different NLG tasks.

    It is important for the research community to collaborate more closely to standardize the evaluation metrics for NLG tasks that are pursued by many research teams. One effective way to achieve this is to organize challenges or shared tasks, such as 
    the Evaluating Natural Language Generation Challenge\footnote{https://framalistes.org/sympa/info/eval.gen.chal} and 
    the Shared Task on NLG Evaluation\footnote{https://github.com/evanmiltenburg/Shared-task-on-NLG-Evaluation}.

    \item \textbf{Developing effective human evaluations.} For most NLG tasks, there is little consensus on how human evaluations should be conducted. Furthermore, papers often leave out important details on how the human evaluations were run, such as who the evaluators are and how many people evaluated the text \citep{van-der-lee-etal-2019-best}. Clear reporting of human evaluations is very important, especially for replicability purposes. 
    
    We encourage NLG researchers to design their human evaluations carefully, paying attention to best practices described in NLG and crowdsourcing research, and to include the details of the studies and data collected from human evaluations, where possible, in their papers. This will allow new research to be consistent with previous work and enable more direct comparisons between NLG results. Human evaluation-based shared tasks and evaluation platforms can also provide evaluation consistency and help researchers directly compare how people perceive and interact with different NLG systems.

    \item \textbf{Evaluating ethical issues.}
    There is still a lack of systematic methods for evaluating how effectively an NLG system can avoid generating improper or offensive language. The problem is particularly challenging when the NLG system is based on neural language models whose output is not always predictable. As a result, many social chatbots, such as XiaoIce \citep{zhou2020design}, resort to hand-crafted policies and editorial responses to make the system’s behavior predictable. However, as pointed out by \cite{zhou2020design}, even a completely deterministic function can lead to unpredictable behavior. For example, a simple answer “Yes” could be perceived as offensive in a given context. 
    For these reasons and others, NLG evaluations should also consider the ethical implications of their potential responses and applications.
    %AI ethics which also covers algorithmic bias is the key to evaluating the generated content to prevent AI-enabled fake news, and video generation (deepfakes), and many others. 
    We should also note that the landscape and focus of ethics in AI in general is constantly changing due to new advances in neural text generation, and as such, continuing development of ethical evaluations of the machine-generated content is crucial for new advances in the field.  
    %\jianfeng{Another important problem is how to evaluate the social impact of NLG systems. This is related to ethics concerns, AI for good, etc. For example, see sec. 7 in \cite{zhou2020design}.}
\end{itemize}

We encourage researchers working in NLG and NLG evaluation to focus on these challenges moving forward, as they will help sustain and broaden the progress we have seen in NLG so far.

\bibliography{style/iclr2019_conference}
\bibliographystyle{style/iclr2019_conference}
\appendix
\end{document}